%% file: LUCSS.tex
\documentclass[acmtog]{acmart}
\setcopyright{acmcopyright}

\setcitestyle{square}

\usepackage{booktabs} 
\usepackage{color}
\usepackage{float}
\usepackage{alltt}
\usepackage{amsmath}
\usepackage{newlfont} 
\usepackage{fixltx2e}
\usepackage{subfig} 
\usepackage{multirow}
\usepackage{amssymb}
\usepackage{enumitem}
\usepackage{wrapfig}
\usepackage{afterpage}
\usepackage{listings}

\usepackage[linesnumbered,ruled,vlined]{algorithm2e}
\usepackage{algpseudocode}  
\usepackage{amsmath}

\begin{document}

\title{LUCSS: Language-based User-customized Colorization of Scene Sketches}

\author{Changqing Zou}
\authornote{Both authors contributed equally to the paper}
\affiliation{\institution{University of Maryland, College Park, United States}}
\author{Haoran Mo}
\authornotemark[1]
\affiliation{\institution{Sun Yat-sen University, China}}

\author{Ruofei Du}
\affiliation{\institution{University of Maryland, College Park, United States}}

\author{Xing Wu}
\affiliation{\institution{Sun Yat-sen University, China}}

\author{Chengying Gao}
\affiliation{\institution{Sun Yat-sen University, China}}

\author{Hongbo Fu}
\affiliation{\institution{City University of Hong Kong, Hongkong}}
\authorsaddresses{}

\renewcommand{\shortauthors}{Zou et al.}

\input{0_abs}



\begin{teaserfigure}
  \includegraphics[width=\linewidth]{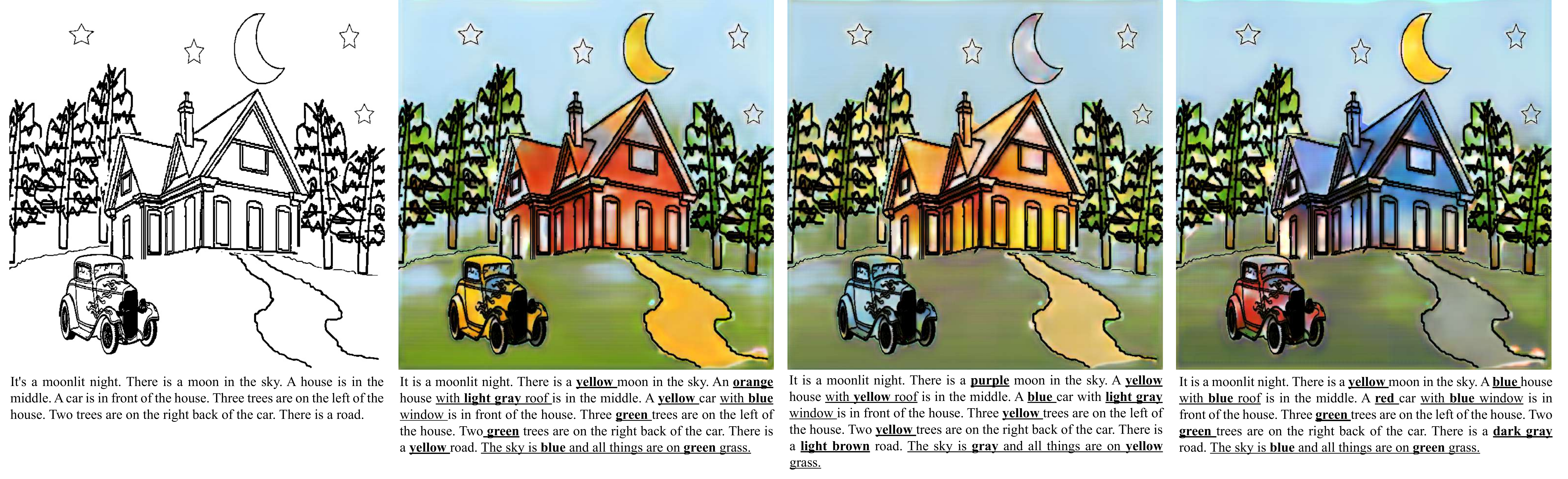}
  \caption{
 {We present LUCSS, a language-based interactive colorization system for scene sketches.  It takes advantage of both instance level segmentation and language models in a unified generative adversarial network, allowing users to accomplish different colorization goals in a form of language instructions. Left: input scene sketch and its content description  automatically generated by LUCSS. Three right columns show the colorization results generated by LUCSS following three different instructions at the bottom. Texts underlined are user-specified, with the target colors highlighted in bold. }
  }
\label{fig:teaser}
\end{teaserfigure}

\maketitle
\input{0_abs}
\input{1_intro}

\input{2_related}
\input{3_overview}

\input{4_segment}
\input{4_segment_captioning}

\input{5_colorization}

\input{6_result}
\input{6_result_colorization}
\input{6_result_object}

\input{6_result_background}
\input{6_result_userscore}
\input{7_discuss}

\bibliographystyle{ACM-Reference-Format}
\bibliography{LUCSS}

\end{document}

%% file: 0_abs.tex
\begin{abstract}
\textbf{Abstract.} 
We introduce LUCSS, a language-based system for interactive colorization of scene sketches, based on their semantic understanding. 
LUCSS is built upon deep neural networks trained via a large-scale repository of scene sketches and cartoon-style color images with text descriptions. 
It consists of three sequential modules. 
First, given a scene sketch, the segmentation module automatically partitions an input sketch into individual object instances. 
Next, the captioning module generates the text description with spatial relationships based on the instance-level segmentation results. 
Finally, the interactive colorization module allows users to edit the caption and produce colored images based on the altered caption.
Our experiments show the effectiveness of our approach and the desirability of its components to alternative choices.

\end{abstract}

%% file: 1_intro.tex
\section{Introduction}
\label{sec:intro}
Sketching is one of the most efficient and compelling ways to communicate complex ideas among humans. 
While abstract sketches can be easily understood by us, teaching machines to understand the underlying semantics of sketches remains a challenging task. 
Recent research has achieved semantic understanding for individually-sketched objects and fostered applications such as sketch-based shape retrieval~\cite{WangKL15} and sketch classification~\cite{eitz2012humans}. 

Nevertheless, instance-level understanding, as often used in scene sketches, has not received much attention. 
Scene sketches usually contain multiple sketched objects, depict scenes of real or imaginary worlds, and widely appear in various scenarios, such as story books, sketch movies, and Computer-Aided Design (CAD) software. 
In this paper, we investigate the instance-level segmentation and interpretation of scene sketches. 
Our work can benefit a number of applications such as sketch-based co-retrieval~\cite{XuCFSH13} and context-based sketch classification \cite{Zhang2018}, which currently take as input manual segmentation of individual scene objects.

With recent advances of deep neural networks and the availability of large-scale datasets such as  MS COCO~\cite{LinMBHPRDZ14} and ImageNet~\cite{Deng2009ImageNet}, machines have outperformed humans in various image understanding tasks for natural images, such as image classification, face recognition, and image generation. Nevertheless, the power of deep networks for scene sketches is still unexplored, and the applications based on scene sketch understanding have rarely been investigated. 

To address these problems, in~\cite{ZouSketchyScene}, we have built a large-scale scene sketch dataset, called \emph{SketchyScene}, and conducted initial studies on 
category-level semantic segmentation of scene sketches. 
In this paper, we investigate two research goals: 1) the capability of deep neural networks for scene sketch understanding, and 2) the potential applications based upon \emph{instance-level} understanding.

To achieve these two goals in a unified framework, we present LUCSS, a language-based interactive colorization system, which consists of three interrelated modules: instance segmentation, captioning, and colorization. 
The instance segmentation module addresses how to employ deep networks for segmenting an input scene sketch into object instances. 
The colorization module can be considered as an application which produces a colorized image conforming to user-specified color requirements for segmented objects. 
Serving as a link between the segmentation and colorization modules, the captioning module takes the output of the segmentation module as input, and automatically generates a caption describing the input scene sketch. Motivated by the recent success of intelligent personal assistants such as \emph{Apple Siri} and \emph{Amazon Alexa}, which are enabled by speech recognition and natural language processing, we present a language-based approach that enables users to embed customized colors into the text description, instead of drawing color strokes on the sketch to specify the colors.
Our approach is more compatible with voice commands for future multimodal colorization systems that can further enhance the user experience.


In this paper, we mainly address two challenging research problems. 
First, how shall we achieve precise instance-level segmentation?
High-quality segmentation results are crucial to the subsequent colorization process, since users might specify different colors for individual objects. 
Unlike natural images, an input scene sketch consists of merely black lines and white background. 
Inferring instance segmentation of a sketched scene is thus challenging due to the sparsity of the visual features (for example, the foreground pixels only occupy 13\% of all the pixels in \emph{SketchyScene} \cite{ZouSketchyScene}). { Employing the state-of-the-art segmentation methods designed for natural images (e.g., \cite{HeGDG17})  directly on sketches does not provide promising results, as shown in Section \ref{sec:segresults}.}
To address this problem, we enhance powerful segmentation models designed for natural image segmentation with the unique characteristics of scene sketches. 

Second, how shall we colorize a high-resolution scene sketch with respect to language-based color inputs? 
Although the colorization of a single sketched object has been extensively investigated, 
the colorization of scene sketches with customized color labels remains an open problem. 
This challenge requires our system to build accurate correspondences between the object instances (parts of object instances) and text-based color specifications. 
Additionally, it requires the system to infer both object-level and object-part-level segmentation.
For example, a user may assign different colorization goals to the window of the car as in shown in Figure~\ref{fig:teaser}.
To tackle this problem, we present a novel architecture which embeds LSTM (Long Short-Term Memory) to a sophisticated Generative Adversarial Network (GAN). 

Furthermore, generating high-quality and high-resolution colorization results ($768\times768$)  is not a trivial task. 
We address this issue by using a two-stage pipeline consisting of object colorization and background colorization. 
Our experimental results (Section~\ref{sec:colorizationresults} and the supplementary materials) show that the LUCSS colorization framework achieves visually pleasing results.

We highlight the major contributions of LUCSS~as follows: 
\begin{enumerate}
\item The first language-based, user-customizable colorization framework for scene sketches. 
\item The first solution for instance-level segmentation  of scene sketches.
\item A colorization dataset of scene sketches with text description and instance-level segmentation. 

\end{enumerate}

%% file: 2_related.tex
\section{ Related Work}
\label{sec:related}
\newcommand{\etal}{{\em et al.}} 
Our work is inspired by and build upon previous work in image segmentation, colorization, captioning, and generation with convolutional neural networks (CNNs) and conditional generative adversarial networks (cGANs).

\subsection{Image Segmentation}
In recent years, CNNs have been proven to yield the state-of-the-art accuracy in semantic object segmentation~\cite{ShelhamerLD16,DeepLabV3,ZhaoSQWJ16,CP2016Deeplab}.
The success of these methods is driven by large-scale, manually-annotated datasets, such as ImageNet~\cite{Deng2009ImageNet} and MS COCO~\cite{LinMBHPRDZ14}, which consist of millions of photographs with segmented objects. 
These methods often jointly predict a segmentation mask and an objectness score based on some appearance features that are specific to an individual class. 
Instance segmentation has become more common after the introduction of the R-CNN pipeline using category-independent region proposals~\cite{HariharanAGM14}. 
Recently, a few methods have shown promise in the task of end-to-end trained instance segmentation~\cite{Girshick15,renNIPS15fasterrcnn,HeGDG17}. 
These approaches perform local and spatially-varying  ``objectness'' estimates, with a simple global aggregation step in the end. 

In our prior work~\cite{ZouSketchyScene}, we conducted a pilot study on category-level semantic segmentation of scene sketches.
In this paper, we investigate the problem of instance-level segmentation in scene sketches. Existing solutions for instance segmentation of natural images (e.g.,~\cite{HeGDG17}) cannot produce satisfactory results for scene sketches because they do not consider the unique characteristics of scene sketches.
Please refer to Section~\ref{sec:segmentation} for further discussion.

\subsection{Sketch Understanding} 
Sketch recognition is perhaps the most popular problem in sketch understanding. 
Since the debut of TU-Berlin dataset~\cite{eitz2012hdhso}, many approaches have been proposed and the state-of-the-art approaches have even outperformed human beings in terms of the recognition accuracy~\cite{YuYLSXH17}. 
Prior algorithms can be roughly classified into two categories: 
1) those using hand-crafted features~\cite{eitz2012hdhso,Schneider2014}, and 
2) those learning deep feature representations~\cite{YuYLSXH17,ha2017neural}.
The latter generally outperform the former by a clear margin. 

Another stream of work has delved into parsing sketched objects into their semantic parts. 
Sun~{\em et al.}~\shortcite{SunWZZ12} proposed an entropy descent stroke merging algorithm for both part-level and object-level sketch segmentation. 
Huang~{\em et al.}~\shortcite{HuangFL14} leveraged a repository of 3D template models composed of semantically segmented and labeled components to derive part-level structures. 
Schneider and Tuytelaars~\shortcite{Schneider2016} performed sketch segmentation by looking at salient geometrical features (such as T-junctions and X-junctions) via a Conditional Random Field (CRF) framework. 
Instead of studying single object recognition or part-level sketch segmentation, we conduct an exploratory study for scene-level parsing of sketches, by using the large-scale scene sketch dataset SketchyScene~\cite{ZouSketchyScene}.

\begin{figure*}[tb]
\centering
   \includegraphics[width=0.95\linewidth]{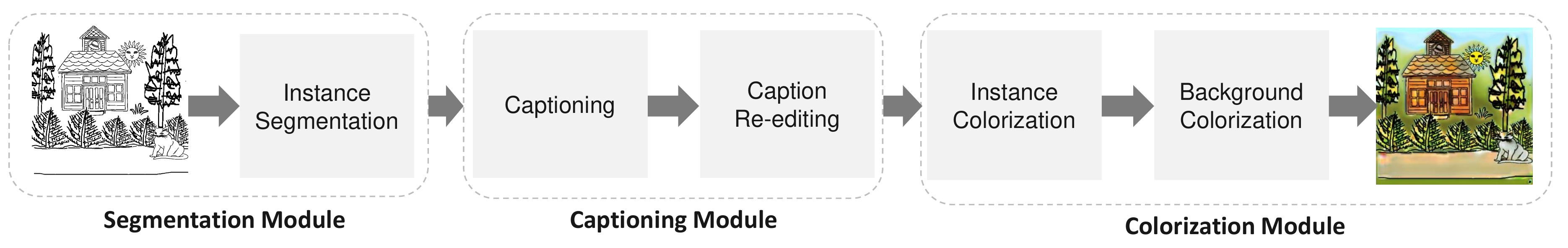}
  \vspace{-7pt}   
   \caption{The modules and the workflow of the LUCSS pipeline.}
\label{fig:overview}
  \vspace{-7pt}   
\end{figure*}

\subsection{Image Captioning}
With the recent advances in  deep neural networks and language learning models, a number of impressive algorithms for image captioning have been developed~\cite{DonahueHGRVSD14,MaoDeepCaption2014,ChenZ14a,XuBKCCSZB15,VinyalsTBE16,zhou2016image}.
We direct readers to a recent survey~\cite{bernardi2016automatic}, which summarizes related datasets and an evaluation of various models.
More recently, attention mechanisms have been applied to the network for narrowing down subjects and thus improving captioning results~\cite{XuBKCCSZB15,DasAZPB17,Anderson2017up-down}.
These modern image captioning models generally consist of two parts: an image encoder and a language model.
The image encoder encodes a raw image into a feature map using a CNN, 
while the language model generates text sequentially with the extracted feature map and the inherited probabilistic dependency. 
In contrast to prior works which aim to generate human-like extractive captions, our captioning module directly produces a lower-level detailed description, which covers the entire scene sketch based on the results of instance segmentation. 
We use this instance-segmentation-based captioning module for user input assistance. 

\subsection{Scene Sketch Based Applications}
While there is little work on semantic segmentation of scene sketches,
some interesting applications have been proposed to utilize scenes with pre-tagged or pre-segmented sketched objects as input. 
For example, Sketch2Photo~\cite{ChenCTSH09} combines sketching and photo montage for realistic image synthesis. Sketch2Cartoon~\cite{sketch2cartoon} is a similar system which focuses on cartoon images.
Similarly, Xu~{\em et al.}~\shortcite{XuCFSH13} propose Sketch2Scene, an automatic system which generates 3D scenes by co-retrieving and co-placing 3D shapes with respect to a scene of pre-segmented sketched objects. 
Sketch2Tag~\cite{SunWZZ12a} is a sketch-based image retrieval (SBIR) system, where scene items are automatically recognized and used as a text query to improve the retrieval performance. 
Our work provides automatic instance segmentation algorithms for scene sketches, and can immediately benefit the above applications.  

\subsection{Image Colorization}
Image colorization assigns a three-dimensional label (RGB) to each pixel from an input gray-scale or sketch image. 
Early studies of colorization methods are mainly based on user interaction~\cite{Huang2005AED,Qu2006,Yatziv2006} or similar examples~\cite{Charpiat2008AIC,Welsh2002TCG} from gray-scale photographs.
To achieve user-customized colorization, interactive strokes~\cite{Huang2005AED,Levin2004} are widely used to provide local guidance based on local intensity differences and spatial offsets~\cite{luan2007natural,qu2006manga}.
Further approaches using local guidance devise better similarity metrics by employing long range connections~\cite{an2010user,xu2009efficient} and local linear embeddings~\cite{chen2005local} to minimize user efforts.
In addition to local guidance, non-local mean-based patch weight~\cite{yao2010patch} and color palette~\cite{chang2015palette} have also been proposed to provide global guidance. 

Recent colorization systems~\cite{Cheng2015DeepColorization, Deshpande2015,Iizuka2016,LarssonMS16,Yan2016}, take advantage of deep CNNs and large-scale datasets to automatically produce plausible color images from gray-scale inputs. Adapting these models to scene sketches is a challenge since scene sketches are sparser than gray-scale images.   Another line of research focuses on sketch colorization which generates color images from black-and-white sketches~\cite{GucluturkGLG16,LiuQLW17,SangkloyLFYH17,XianSLFYH17}. 
However, most of the prior sketch colorization approaches target object-level sketches, while our work is a scene-level sketch colorization method. 

Sangkloy~\etal~\shortcite{sangkloy2017scribbler} has developed a system to translate sketches to real images, with colorization goals assigned by user color strokes. This system can well convey the user's colorization goals to image regions but its extensibility may be limited.  PaintsChainer~\cite{PaintsChainer2017} and Frans~\etal~\shortcite{frans2017outline} have developed open-source interactive online applications for line-drawing colorization. 
Concurrently, Chen~\etal~\shortcite{Jianbo2018RAM} proposed a language-based colorization method for object-level sketches or gray-scale images. 
Our system is close to~\cite{Jianbo2018RAM} but we focus on scene-level sketch colorization.  

\subsection{Image Generation with GANs}
Several recent studies, such as DCGAN \cite{RadfordMC15}, Pix2Pix \cite{pix2pix2017}, WGAN \cite{arjovsky17a}, and SketchyGAN \cite{sketchGAN}, have demonstrated the value of variant GANs for image generation. The variations of conditional GANs have been further applied to text-to-image synthesis~\cite{pmlrReed16,ZhangXL17,Hong2018}, image inpainting~\cite{OordKK16,pathakCVPR16context,YehCLHD16}, and image super-resolution~\cite{Ledig2017,SonderbyCTSH16}. 
Nonetheless, all the aforementioned generators are conditioned solely on text or images.
In contrast, LUCSS takes both image and text as input, presenting an additional challenge of fusing the features of a scene-level image and the corresponding text description.  

%% file: 3_overview.tex
\section{System Overview}
\label{sec:sys}

As illustrated in Figure~\ref{fig:overview}, our system processes an input sketch through multiple stages. 
It first performs instance segmentation to recognize and locate individual objects in the scene sketch, and then automatically produces the caption describing lower-level detailed information of the sketch ({\em e,g.}, object category, object position, quantity) with a template-based algorithm. Next, the user can specify colorization goals by changing the automatically-generated caption. 
Finally, the system colorizes the whole sketch into a color image, where each segmented instance meets the user's requirements via a novel cGAN based model. 

The system contains three sequential modules. 
The first one is an instance segmentation deep network. 
It takes as input a scene sketch image, and outputs the class labels and the instance identity label for each pixel belonging to the objects in the scene. 
In this module, we adapt the state-of-the-art segmentation models (including Deeplab V2 and Mask-RCNN) to the sketch data via analyzing the characteristics of the sketch data. 
The segmentation accuracy is significantly improved compared to the original models. 

The second module is a language template-based caption generator. 
Taking the segmentation results as input, it analyzes the geometric relationships between object instances (e.g., position and occlusion relationships) and generates object-level scene descriptions. 
Figure~\ref{fig:teaser} shows a typical result of this module. 
Besides captioning, this module provides a friendly interface for the user to assign different editing goals for individual object instances. 
This module can facilitate the system to build the exact correspondence between object instances in the scene and their corresponding sentences (since the caption is generated by the system from the segment instances, correspondences can be easily obtained by comparing the user-modified caption with the original one). 
In contrast to recent research~\cite{Jianbo2018RAM}, which implicitly infers the correspondence between object parts and sentences using an attention mechanism, we leverage the captioning module to achieve more accurate correspondence, especially for complex scenes. As shown in the experiment of Section~\ref{subsec:mainRes}, implicit inference usually fails on a long caption with more than six sentences, but our approach has no such constraint. Hence it provides a better assurance that  results will achieve user specifications. 
%
%

The last module is a typical application based on the instance segmentation results. In this work we focus on the colorization task. Unlike recent research which aims to colorize single objects or relatively simple scenes~\cite{autoPainter2017, Varga2017,Jianbo2018RAM}, our work aims to solve a complex scene sketch colorization problem with the help of language-based user instruction. Our experiment in Section \ref{sec:colorization} found that even the most best model to date ~\cite{Jianbo2018RAM} is unable to successfully colorize each individual object instance to exactly meet the user's needs. The main reason is that a user's language instructions usually have ambiguity, especially when the target scene is complex (e.g., there are multiple object instances belonging to the same category in a scene). 
Our solution is to decompose the colorization of the whole scene sketch into two sequential steps: instance colorization and background colorization. In instance colorization, each object instance is colorized. While in background colorization, the remaining regions which do not belong to any object instance are colorized. This divide-and-conquer strategy leverages the natural advantages of the segmentation approach, making the colorization of a complex scene sketch resolvable.   

\textbf{Dataset.}
We use SketchyScene as the basic dataset to study the segmentation and colorization problem. SketchyScene contains 7,264 scene templates, each scene template corresponding to a color reference image. Moreover, SketchyScene provides the ground-truth for both semantic segmentation and instance segmentation of  $7,264 $ scene sketches. Figure \ref{fig:SketchyScene} shows a typical example of a scene sketch in SketchyScene. See the supplementary material for more details of SketchyScene.  
%
%

%% file: 4_segment.tex
\section{segmentation and captioning}
\label{sec:segmentation}

\begin{figure}[]
\centering
   \includegraphics[width=0.95\linewidth]{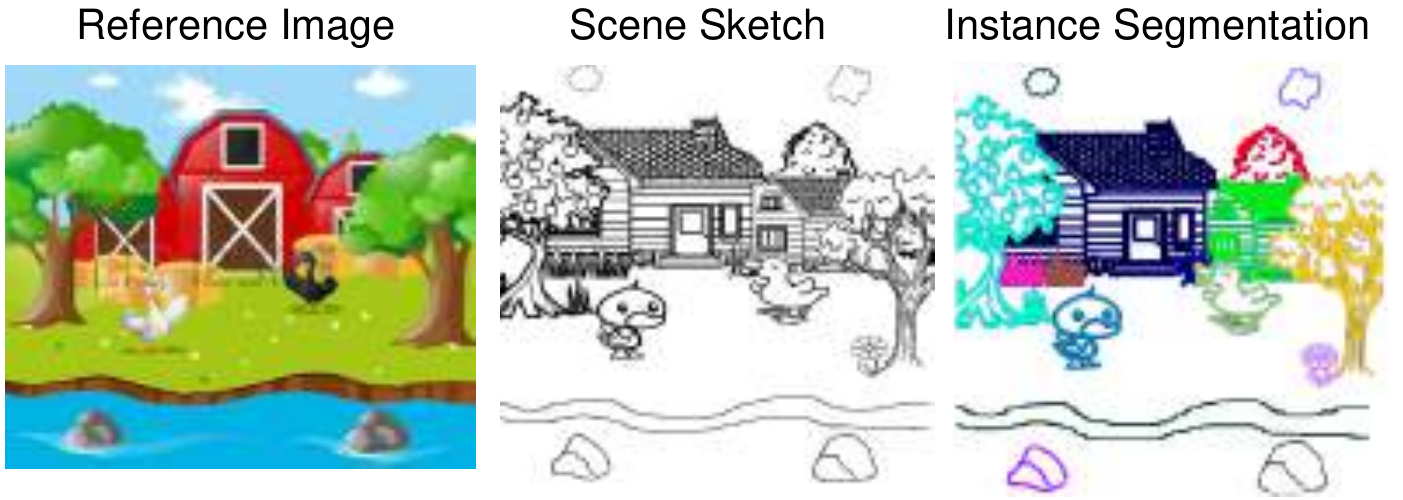}
  \vspace{-7pt}   
   \caption{An example scene template from SketchyScene. (a) shows a cartoon-style reference image harvested from the Internet, (b) shows a user-synthesized scene sketch using a repository of instance sketch, and (c) shows the ground-truth labels of the instance segmentation.}
\label{fig:SketchyScene}
  \vspace{-7pt}   
\end{figure}

\subsection{Instance Segmentation}
\textbf{Formulation.} Instance segmentation segments individual object instances, {possibly of the same object class} in a scene sketch. 
This is challenging, especially when the instances of the same class have occlusions, {\em e.g.}, two trees growing together, as shown in Figure \ref{fig:SketchyScene}. 
Apart from using pixel-level class labels as supervision, spatial information like object bounding box is necessary. Generally, in an instance-level segmentation task, each unknown instance $I$ in the input image can be denoted as a tuple $[B, L, M]$. 
Here, $M$ is a binary mask covering the instance, $L$ is a class label, and $B$ is the 4-D vector encoding the position and size of the bounding box in the format of $[x, y, H, W]$, where $[x, y]$ indicates the top-left corner and $H$ and $W$ represent the height and width, respectively. {Our goal of instance segmentation is to assign pixels of a scene sketch image to a specific instance mask $M$ located in an inferred $B$}. 

Unlike natural images, a sketch only consists of black lines and a white background. 
Given that only black pixels convey semantic information, our problem for instance segmentation of a scene sketch is defined as predicting $[B, L, M]$ for each black pixel. Taking the rightmost image of Figure \ref{fig:SketchyScene} as an example, when segmenting trees, duck, house and cloud, every black pixel should be assigned to a specific instance mask $M$ with a class label $L$, located in $B$, while the remaining white pixels are treated as background.

\begin{figure}[]
\centering
   \includegraphics[width=0.99\linewidth]{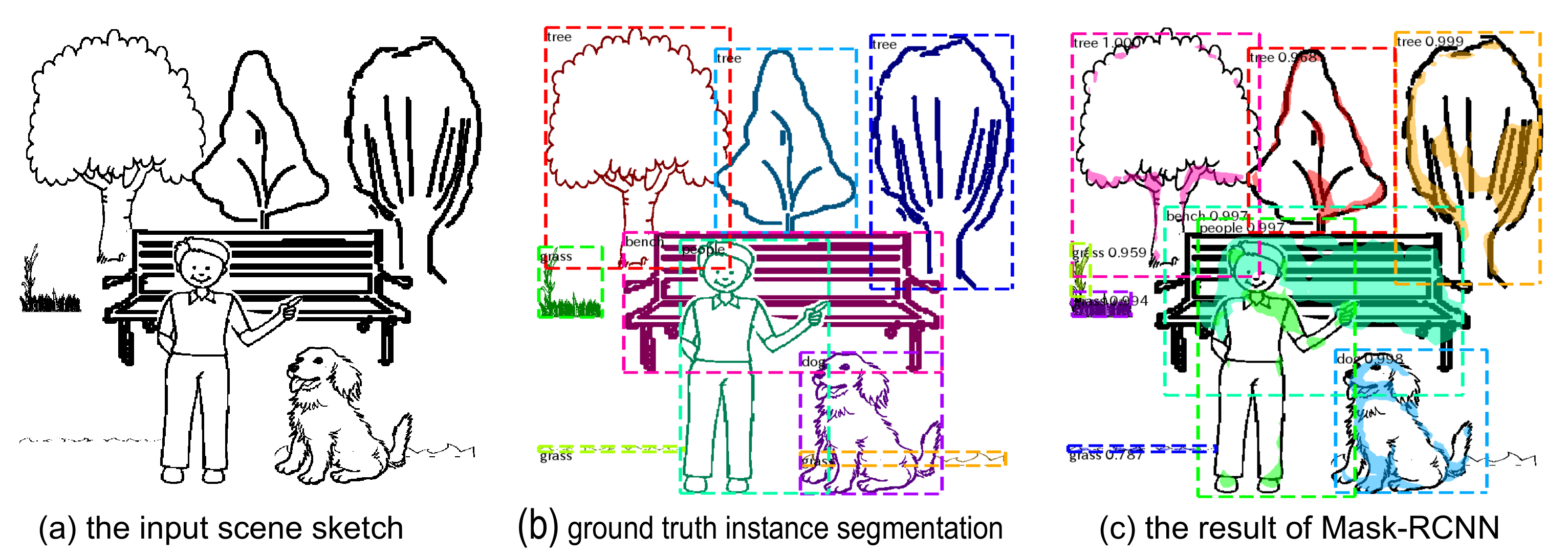}
  \vspace{-7pt}   
   \caption{An illustration of instance segmentation using the state-of-the-art model Mask-RCNN \protect\cite{HeGDG17} on a representative scene sketch of SketchyScene. Note that though the bounding boxes in (c) from the Mask-RCNN model are roughly aligned with the ground truth labels (b), the Mask-RCNN model misclassifies the strokes of the person as the bench class and fails to follow the black lines at the pixel level.
}
\label{fig:Mask-CNN}
  \vspace{-7pt}   
\end{figure}

\textbf{Challenges.}
Segmenting a sketchy scene is challenging mainly due to the sparsity of visual features. First, a scene sketch  image is dominated by white pixels. For the $7,264$ examples of SketchyScene, the average background ratio is $87.83$\%. The remaining pixels belong to foreground classes. The classes, as measured by their sketch pixels, are thus quite imbalanced. Second, segmenting occluded objects in sketches is much harder than in natural images, where an object instance often contains uniform colors or texture so that context information can help in segmentation. Unfortunately, such cues do not exist in a scene sketch.

The sparsity of visual features causes the instance segmentation models, designed for natural images, perform poorly on scene sketches. 
In an initial experiment, we found that the segmentation results are unsatisfactory even with the state-of-the-art model Mask-RCNN \cite{HeGDG17} (see a representative result in Fig. \ref{fig:Mask-CNN} (3)).  Although Mask-RCNN can successfully detect the bounding boxes of object instances in a scene sketch, the binary masks often fall out of the black lines. 

\textbf{Methodology.}
Our initial study in \cite{ZouSketchyScene} has reported a significant finding on a semantic segmentation task. 
That is, the challenge for sketch scene segmentation is mainly caused by the large area of background. 
In a model tailored to this property the background pixels should not contribute to the loss during training. During the inference, background pixels are assigned to an arbitrary class label, and are filtered out by the drawing mask of the input sketch for the final output. Using this strategy, the adapted DeepLab-v2 model \cite{ZouSketchyScene} can improve the semantic segmentation MIoU (Mean Intersection over Union) by more than $10\%$.  On the test examples, this model achieved the best performance ($63.1\%$ on MIoU) over DeepLab-V3 models~\cite{DeepLabV3}, SegNet~\cite{badrinarayanan2017segnet}, and
FCN-based~\cite{long2015fully} model.  Inspired by this finding, we have also tailored Mask-RCNN with this property, {\em i.e.}, ignoring the contribution of 
the background pixels when training the model. 
This adapted model also significantly improves the segmentation quality (Section~\ref{sec:segresults}). 

Inspired by recent methods~\cite{ArnabT17,LiangWSYLY15}, which adopt a segmentation-first strategy, our final solution for instance segmentation is a framework which infers $B$ and $M$ based on semantic segmentation results. 
Specifically, we use Mask-RCNN to generate the bounding boxes $B$  and $L$.
We use the semantic segmentation results generated by an improved variation of the DeepLab-v2 model within the region of $B$ to compute $M$.   

\begin{figure}[]
\centering
   \includegraphics[width=0.95\linewidth]{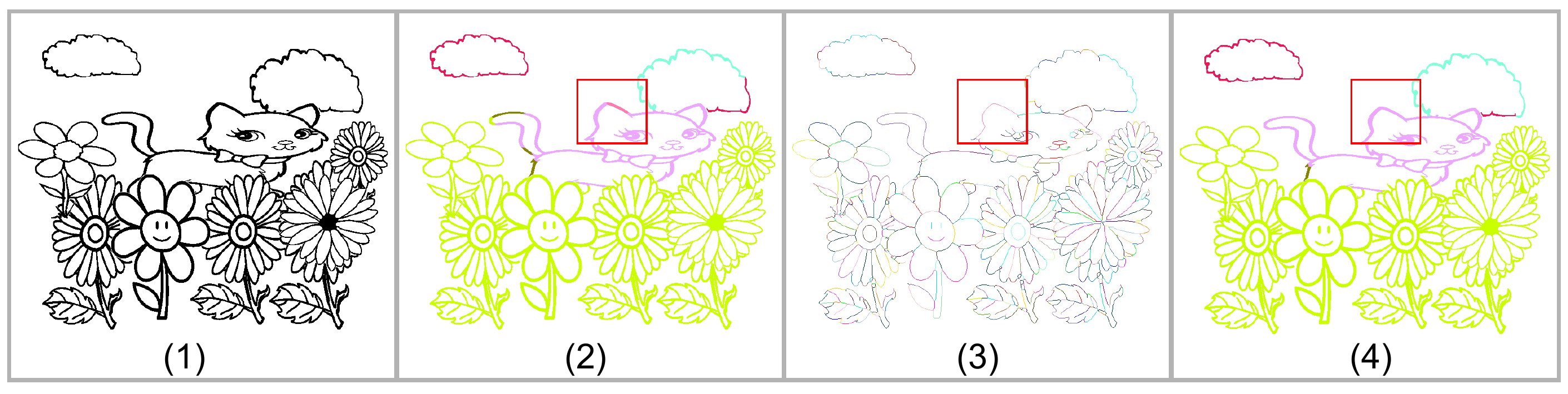}
  \vspace{-7pt}   
   \caption{Result of the adapted DeepLab-v2 model can be further refined by a postprocess based on edgelist, in which each entity is a skeleton of a drawing cluster. From left to right: (1) scene sketch, (2) output of the adapted DeepLab-v2 model in \cite{ZouSketchyScene}, (3) edgelist, and (4) post-processed result with edgelist.}
\label{fig:edgelist}
  \vspace{-7pt}   
\end{figure}

Although the DeepLab-v2 model leverages a densely connected conditional random field (CRF) as the post-processor to refine the label inference, 
it still produces some discontinuous labels even on drawings which appear to belong to the same ``edge" (see the ear of the cat in Figure~\ref{fig:edgelist}). 
The degradation of CRF on sketches is mainly caused by the sparsity of sketches. 
CRF models the similarity of the class labels in an 8-connected neighborhood, while 
in a sketch a non-background pixel is usually affected by the neighboring pixels along sketch lines. 
This motivated us to use an assumption based on \textit{edgelist} to further improve the results: 
the non-background pixels corresponding to the same drawing cluster (i.e., super-pixel consisting of only black pixels) should have the same class label. 

An edgelist is defined as a set of single-pixel-width edges, in which each edge is a single-pixel-width skeleton of a drawing cluster as shown in Figure~\ref{fig:edgelist}(3). 
Given a sketch, we first extract the single-pixel-width skeleton using the edge-link algorithm \cite{KovesiMATLABCode}. 
After that, we compute the nearest skeleton ({\em i.e.}, an item of edgelist) for each non-background pixel and separate the sketch into a set of drawing clusters. 
%
%
We finally assign the majority class label of a drawing cluster to all the non-background pixels of the same cluster.  
We present the experimental results of this improved framework in Section~\ref{sec:segresults}.  

\begin{figure*}[tbp]
\centering
   \includegraphics[width=0.99\linewidth]{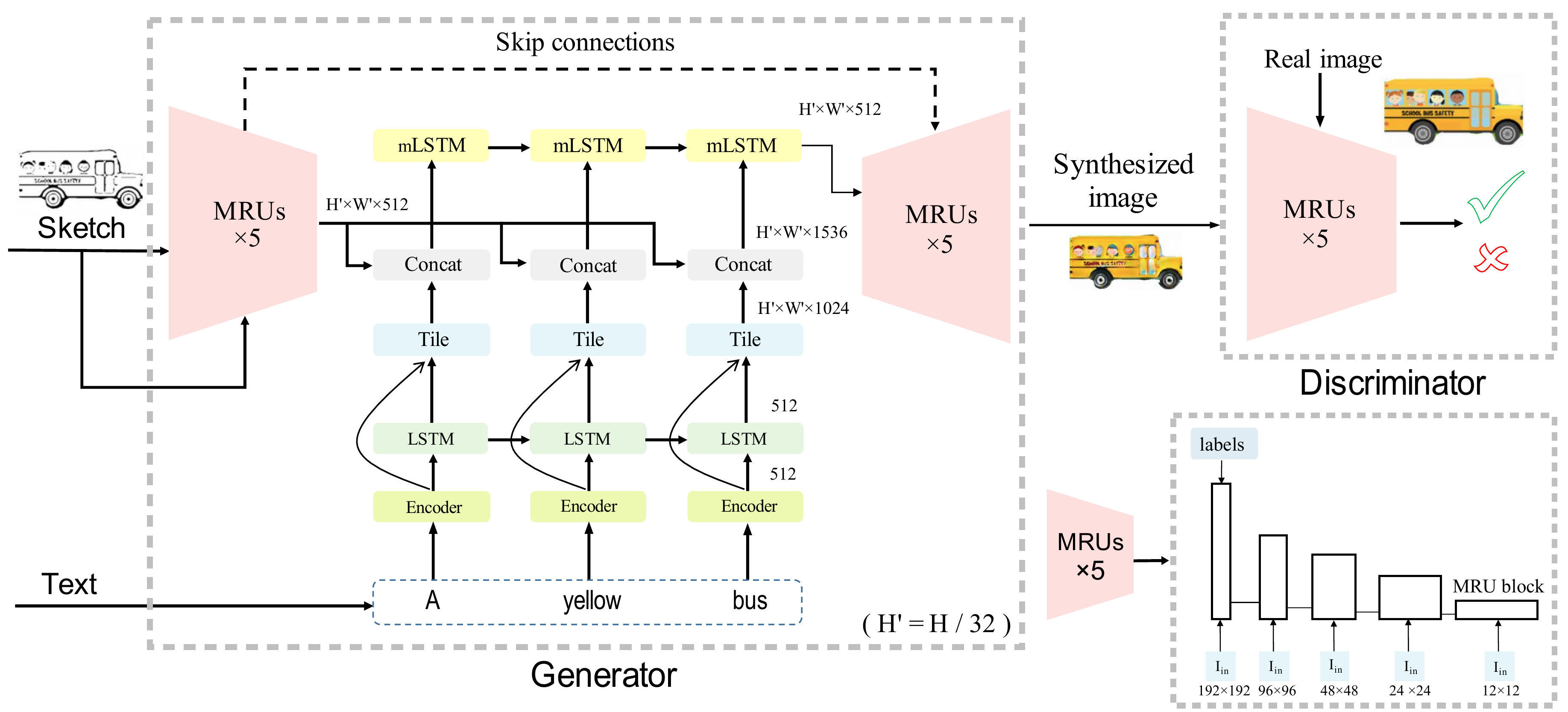}
  \vspace{-7pt}   
        \caption{Network architecture for object colorization, composed of a convolutional image encoder built on MRU blocks, a fusion module consisting of LSTM text encoders and multimodal LSTMs (mLSTM), a de-convolutional image decoder, and a MRU-blocks-based convolutional discriminator. The structure of the MRUs, which is shown at the right bottom corner, is inspired by SketchyGAN \cite{sketchGAN}. 
}
\label{fig:instanceNetwork}
  \vspace{-7pt}   
\end{figure*}

%% file: 4_segment_captioning.tex
\subsection{Captioning}
We develop a template-based algorithm for caption generation that starts from the output of the segmentation module ({\em i.e.}, a list of $[B,L,M]$, where $B_i$, $L_i$, and $M_i$ contain the information of size, location, and class label of an object instance $i$). 
The steps of the algorithm are shown in Algorithm \ref{alg:captioning}. 
It first classifies all the objects into three sets according to their class labels (we make this classification because SketchyScene was built by three sets of objects: objects related to weather or time, objects related to the site environment,  and other objects related to the site environment).
The first set $\textbf{W}$ is for objects related to weather or time ({\em e.g.}, sunny or cloudy, night or day). 
The second set {$\textbf{E}$} is for objects which usually have a larger size and might determine the site environment of the scene sketch ({\em e.g.}, house, road, table, and bus). 
The last set is for the other objects $\textbf{O}$. 
Afterwards, we sequentially describe the objects from [$\textbf{W}, \textbf{E}, \textbf{O}$]. 
For objects from $\textbf{W}$, the algorithm generates a general weather description according to object labels, {\em e.g.}, ``it is a sunny day''. 
For objects from $\textbf{E}$, it produces the sentences which describe the information of class labels and absolute locations. 
A typical description is ``There is a house in the center of the image". 
For objects from  $\textbf{O}$, it produces the description of the formation of class labels and relative locations. 
A typical description is ``A person is in front of the house".

\begin{algorithm}  
	\caption{Sketch Captioning}  
	\label{alg:captioning}
	\KwIn{pred\_boxes $B$, pred\_class\_labels $L$, pred\_class\_masks $M$,}  
	\KwOut{caption $T$}  
	$items$ \textleftarrow Node($B$,$L$,$M$) \
	
	\For{$item$ $\in$ $items$ }  
	{     
		\If{$item$ $\in$ $\textbf{W}$}  
		{  
				$T_{Weather}$\textleftarrow Weather($item$)\ 
		}  
		\If{$item$ $\in$ $\textbf{E}$}  
		{  
			$T_{Environment}$\textleftarrow Environment($item$)\ 
		} 
		\If{$item$ $\in$ $\textbf{O}$}  
		{  
			$T_{Object}$\textleftarrow Object($item$)\ 
		} 				
	}   
	 $T$\textleftarrow $T_{weather}$+$T_{Envirment}$+$T_{Object}$\
	 
	\textbf{return} $T$\
\end{algorithm}

The captioning module is applied to a human-machine interface as shown in the supplementary video. 
It is used to visualize the correspondence between an object in the input sketch and the corresponding sentence in the caption. 
Apart from captioning itself, there are two additional functions: 1) visualize if the sketch parsing results are accurate or not, and 2) visualize if the colorization goal is assigned to a desired object.

%% file: 5_colorization.tex
\section{Colorization}
\label{sec:colorization}
The colorization process contains two sequential steps: object instance colorization and background colorization. 
The object instance colorization step assigns target colors to the pixels belonging to segmented object instances, including their blank inner regions. 
Background colorization assigns colors to the remaining pixels. 

\subsection{Object Instance Colorization }


\textbf{Overview}. The proposed framework for object instance colorization as shown in {Figure \ref{fig:instanceNetwork}}  is a conditional GAN model consisting of a generator $G$ and a discriminator $D$. $G$  takes as input an object sketch image $I$ and its corresponding language description $S = \{w_1,  w_2,...,w_t..., w_T\}$, where $w_t$ are individual words in the sentence, and generates a color image.  Compared with the generators in existing literature \cite{sketchGAN,pix2pix2017}, which learn mappings from an input sketch or image to an output image, our generator $G$ models the interaction among the text description, visual information, and spatial relationships, and finally fuses the multi-modal features together. 
The generation of the color images is controlled by the text information. 
The discriminator, which is the opponent of the generator, is fed with both the generated images and the real color images at the same time, and serves the function of judging whether an image looks real or not.

\textbf{Generator.} The basic architecture of the generator is an encoder-decoder structure built on MRU blocks \cite{sketchGAN}. It consists of three modules: 
an image encoder which encodes
the features of the $H \times W$ input sketch (segmented object sketch), 
a fusion module which fuses the text information into the image feature map generated by image encoder, 
and finally an image decoder which takes the fusion map produced by the fusion module and produces an $H \times W \times C$ map, where $C$ is the number of color channels. 
The MRU block is first proposed in \cite{sketchGAN},  which uses a learned mask to selectively extract features from the input images. 
Its cascaded structure of MRU blocks allows the ConvNet to repeatedly and progressively retrieve the information from the input image on the computation path. 
In this work, we use MRU blocks in both the encoder and decoder.  
In our implementation, we use five cascaded MRUs to encode the $H \times W$ input sketch image into an $H' \times W' $ feature map ($H' = \frac{H}{32}, W' = \frac{W}{32}$) for the encoder. 
For the decoder,  five symmetric MRUs are cascaded as a multi-layer de-convolutional network. 
Skip-connections are applied between the encoder and the decoder, concatenating the output feature maps from the encoder blocks to the output of the corresponding decoder blocks.  

The fusion module fuses the text information in $S$ into the $H' \times W'$ image feature map, and outputs an $H' \times W' $ fusion feature map. 
It is a core module of the generator and inserted into the bottleneck phase of the generator. 
The basic architecture of our fusion module is a convolutional multimodal LSTM, called recurrent multimodal interaction (RMI)  model which 
was used to fuse the information of image referring expressions to segment out a referred region of the image \cite{RMI2017}. 
The typical characteristic of an RMI model is that the language model can access the image from the beginning of the language expression, allowing the modeling of the multi-modal interaction. 
Our work uses RMI to mimic the human image colorization process. 
For each region in the input sketch, the fusion module reads the language feature map repeatedly until sufficient information is collected to colorize the target image region.  

A concurrent system \cite{Jianbo2018RAM} called LBIE (Language-Based Image Editing) has used a similar model to segment and colorize object parts of interest in an image conditioned by language descriptions. We have implemented the fusion module for our generator with both RMI and LBIE. We present the comparison results in the experimental section.

\textbf{Discriminator.} The discriminator $D$ takes in a generated image and outputs the probability of the image being realistic. The structure of the discriminator follows SketchyGAN \cite{sketchGAN} which uses four cascaded MRUs. It takes four types of scales of the real and generated images.

\textbf{Loss and training.} We use a hybrid loss following SketchyGAN \cite{sketchGAN} which includes a GAN loss and an auxiliary classification loss in the Discriminator, a GAN loss, an auxiliary classification loss, an L1 loss, a perceptual loss, and a diversity loss in the Generator. 
The auxiliary classification loss can improve the ability of both the Discriminator and the Generator, and further enhances the quality of the synthesized images. 
The L1 distance loss is for comparison between the synthesized image and the ground-truth cartoon image. 
The perceptual loss and diversity loss are for generating diverse results. 

\subsection{Background Colorization}
\label{sec:bg_color}

The background colorization network takes as input the result of object instance colorization, and infers the colors of the background regions to produce the final colorful image.  The network architecture still employs a cGAN structure similar to that used for object instance colorization, with some modification as detailed below.


\textbf{Generator. } The architecture of the generator is similar to that shown in Figure~\ref{fig:instanceNetwork}. 
We replace MRU blocks with residual blocks \cite{HeZRS16}. 
Both the encoder and decoder use five cascaded residual blocks. 
The residual unit numbers of the five residual blocks for the encoder are \{1, 3, 4, 6, 3\}   (\{3, 6, 4, 3, 1\}  for the decoder). 
We make this change because an MRU block, which uses a binary mask to selectively extract features from a sketch, is more suitable for sketch images than color images. 
Our initial experiments confirm this speculation. 
The colorized backgrounds produced by the generator shown in  Figure \ref{fig:instanceNetwork} ({\em i.e.}, the generator used for object instance)  usually have sharp region boundaries, leading to relatively poor visual effects (also see the results shown in the second column from right of Figure\ref{fig:bgBackbones}).
Apart from the use of residual blocks, we have also conducted some experiments with  the network replacing MRU blocks with the encoder blocks in \cite{pix2pix2017}. 

\begin{figure}[]
\centering
   \includegraphics[width=0.99\linewidth]{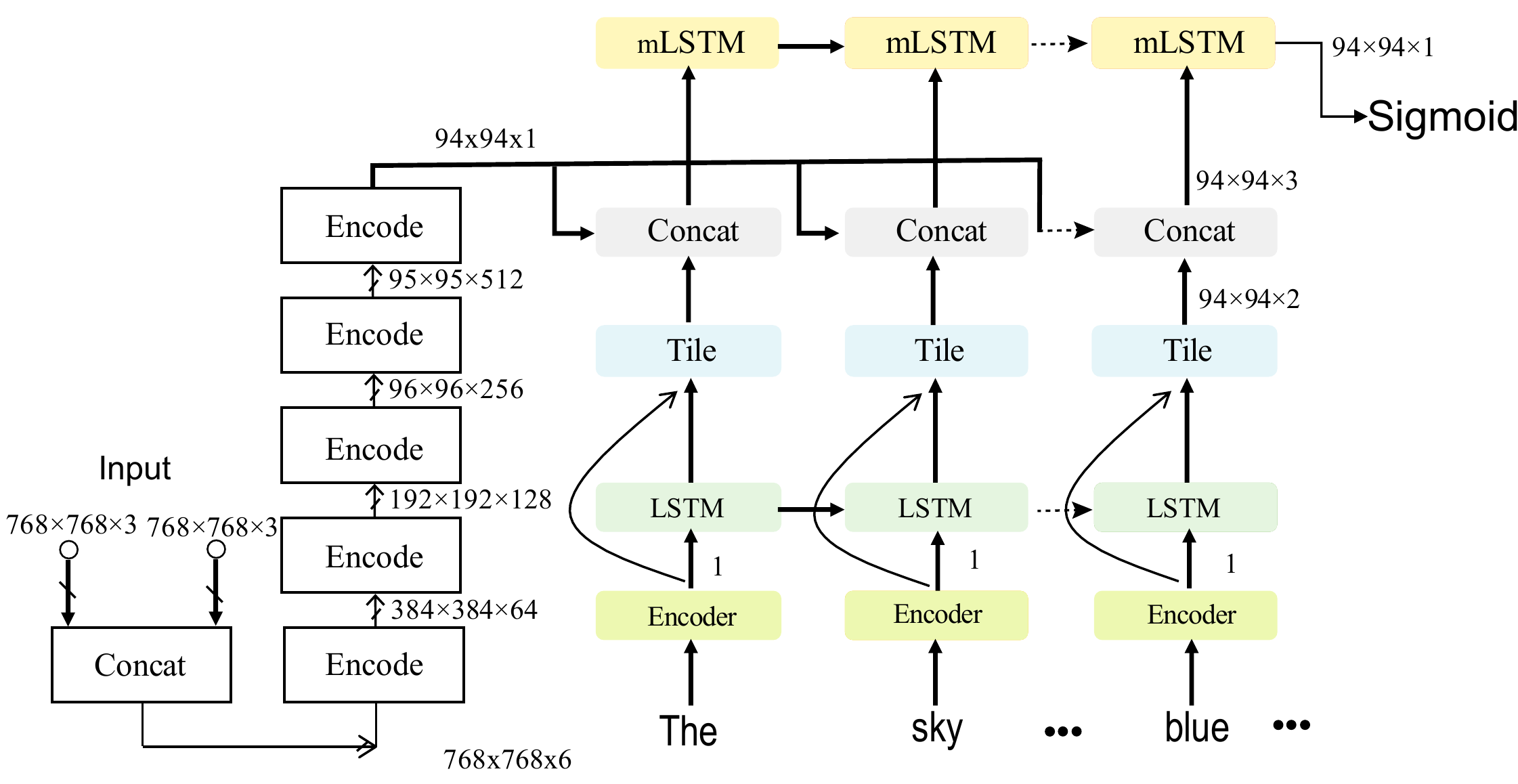}
  \vspace{-7pt}   
        \caption{Structure of the discriminator for background colorization. The structure combines five cascaded encode blocks used in the pix2pix network \cite{pix2pix2017} and RMI model.} 
  \label{fig:bgNetwork}
  \vspace{-7pt}   
\end{figure}

\textbf{Discriminator.} 
For the discriminator, we use the architecture shown in Figure~\ref{fig:bgNetwork}. 
It is a combination of the RMI model and five cascaded encode blocks used in pix2pix \cite{pix2pix2017}. 
Unlike the generator, each of the five cascaded encoder blocks contains a single layer, which decreases the complexity of the entire cGAN. 
In addition, the input text information is also fused into the image feature by the RMI model. 
The joint modeling of the text and image helps the discriminator make a judgment monitored by the text information. 
Our experimental results (shown in supplementary materials) show that the joint modeling improves the capability of the whole network. Note that we do not use the RMI model to fuse the text information for the discriminator of object instance colorization. 
We simplify the discriminator because the discriminator with the fusion module does not improve the colorization significantly on single object sketches with lower resolution ($192 \times 192$) in our initial experiments.

\textbf{Loss and training.} 
As for the loss and training, we follow the scheme of Pix2Pix \cite{pix2pix2017}. 
We only use a conditional GAN loss as well as a L1 distance loss.  The diversity loss
is not used here  as we do for object colorization, because we 
expect to suppress the diversity of the generated background.  

%% file: 6_result.tex
\section{Segmentation Evaluation}
 \label{sec:segresults}
We conducted our experiments for segmentation on SketchyScene. The entire dataset, including 7,264 unique scene sketch templates, was randomly split into training (5,616), validation (535), and test (1,113) datasets.  

\noindent\textbf{Segmentation models.} We compared four types of frameworks: Mask-RCNN (Model-1), Mask-RCNN + w/o BG (Model-2), Mask-RCNN + edgeList (Model-3), and Mask-RCNN + adapted DeepLab-v2 + edgeList (Model-4). These four frameworks are abbreviated as Model-1 to Model-4 below. Model-1 is extended from Faster-RCNN \cite{FasterRCNN2015} by adding a parallel object mask prediction branch and is one of the most advanced methods proposed for instance segmentation on natural images. Model-2 is an adapted Mask-RCNN model, where background pixels do not contribute to the loss during training. Model-3 is extended from Model-2 by adding a post-processing using edgelist. Model-4 is a combination model of the adapted DeepLab-v2 model \cite{ZouSketchyScene} and Model-3, which means only the mask pixels, whose semantic label from adapted DeepLab-v2 is the same with the predicted label from Mask-RCNN, will be retained. The results of all the four models 
were filtered with the binary mask of non-background pixels. Similar to \cite{HeGDG17}, we use $AP$ (MS COCO metric), $AP_{50}$ and $AP_{75}$ (PASCAL VOC metrics) as evaluation metrics.  It is worth mentioning that except for Model-1, all others: Model-2, Model-3, and Model-4 are our methods. The final segmentation model of LUCSS uses Model-3.
%
%

\textbf{Implementation details.} 
We implemented the proposed technique using Deeplab-v2 on Python 3 and TensorFlow, based on a ResNet101 backbone. 
The initial learning rate was set to 0.0001 and mini-batch size to 2. 
We set the maximum training iterations as 50K and the choose SGD (Stochastic Gradient Descent) as the optimizer. 
We resized the images as $768 \times 768$. 
The hyper parameters $\sigma_{\alpha}, \sigma_{\beta}$, and $\sigma_{\gamma}$ of denseCRF in the adapted DeepLab-v2 were set to 7, 3, and 3, respectively. 

\begin{figure}[]
\centering
   \includegraphics[width=0.99\linewidth]{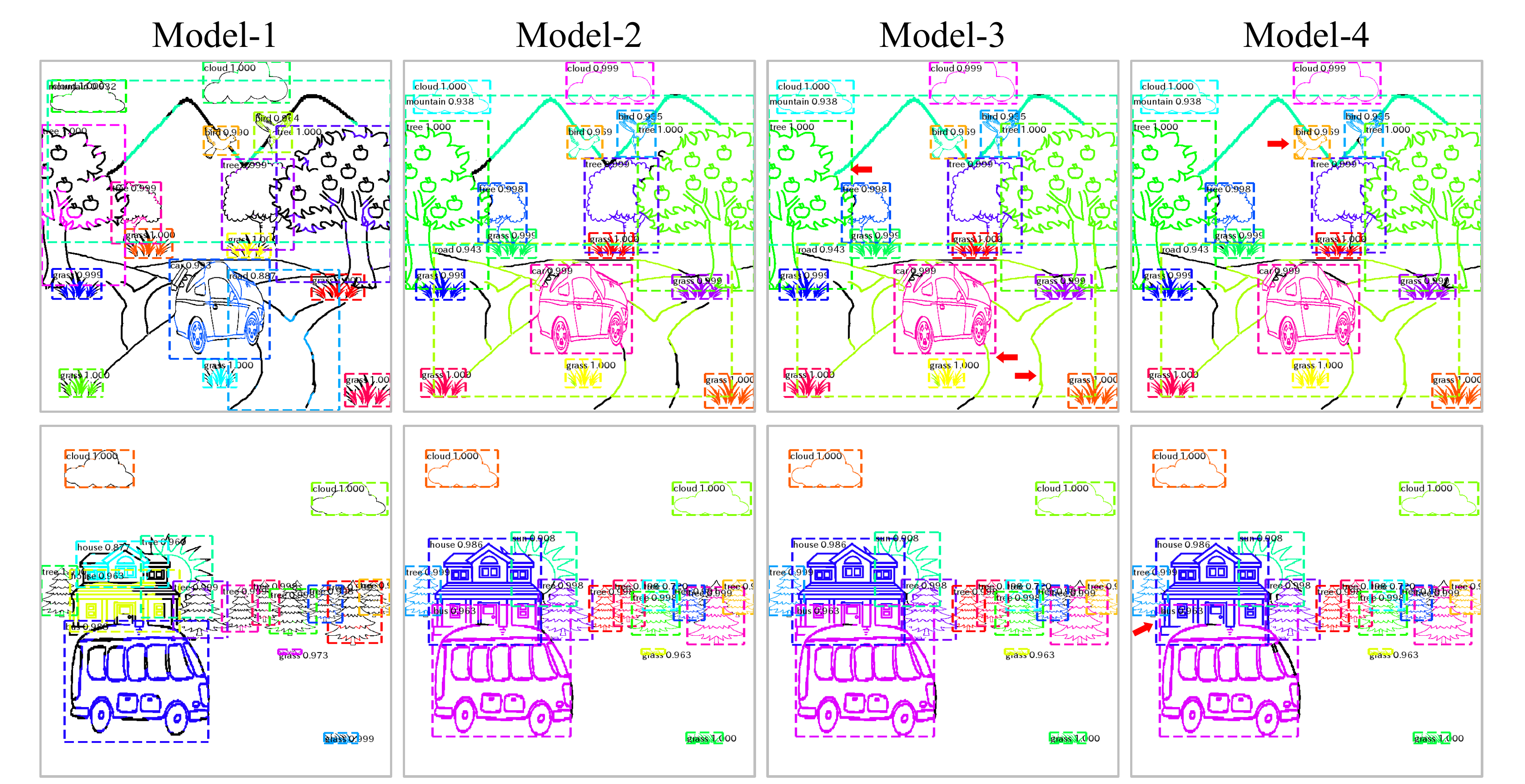}
  \vspace{-7pt}   
        \caption{Qualitative results of four competitors for instance segmentation. The improvement by ignoring background pixels is clear by comparing Model-1 and Model-2. In the third and fourth column, the benefits from edgelist and class labels can be found from the areas where red arrows point to.} 
  \label{fig:visseg}
  \vspace{-7pt}   
\end{figure}

\textbf{Results.} Table \ref{table:baseline} shows the performance of the different models. 
Clearly, Model-2 performs much better than Model-1. 
It indicates that ignoring background pixels improves the segmentation task, which confirms the conclusion we made in \cite{ZouSketchyScene}. 
Although Model-3 achieves slightly higher average quantitative values than Model-2, we find the improvement due to the use of edgelist usually occurs on some non-background pixels, which represent salient category characteristics such as the ear of a cat as shown in Figure \ref{fig:edgelist}. 
The improved performance is thus visually clearer. 
Model-4 obtains slightly better AP value in validation set compared to Model-3, but it gets worse in test set.
Figure~\ref{fig:visseg} shows some segmentation results of the above compared methods. More results can be found in the supplementary materials.

\begin{table}[h]
	\renewcommand\arraystretch{1.2}
	\begin{center}
    \newcommand{\tabincell}[2]{\begin{tabular}{@{}#1@{}}#2\end{tabular}}
		\caption{Instance segmentation mask AP on SketchyScene. Model-2, Model-3, and Model-4 are three models proposed by the work.  
        $AP$ (MS COCO metric), $AP_{50}$ and $AP_{75}$ (PASCAL VOC metrics) are the same metrics as those used in MaskRCNN.}
		\label{table:baseline}

		\begin{tabular}{|c|c|c|c|c|c|c|}
			\hline 
			 \multirow{2}{*}{Model} & \multicolumn{3}{c|}{val} & \multicolumn{3}{c|}{test} \\ 	
	\cline{2-7}
	& AP & $\textrm A\textrm P_{50}$ & $\textrm A\textrm P_{75}$ & AP & $\textrm A\textrm P_{50}$ & $\textrm A\textrm P_{75}$\\ 
			\hline
			Model-1  & 10.03 & 29.20 & 4.86 & 12.99 & 36.76 & 5.93 \\
			\hline
			\tabincell{l}{\textbf{Model-2}}   & 63.01 & 79.97 & 68.18 & 62.32 & 77.15 & 66.76 \\
			\hline
			\tabincell{l}{\textbf{Model-3}} & 63.78 & \textbf{80.19} & 68.88 & \textbf{63.17} & \textbf{77.45} & \textbf{67.60} \\
			\hline
			\tabincell{l}{\textbf{Model-4}} & \textbf{64.38} & 79.24 & \textbf{69.10} & 58.55 & 71.31 & 62.29 \\
			\hline
		\end{tabular}
	\end{center}
\end{table}

%% file: 6_result_colorization.tex
\section{Colorization Evaluation}\label{sec:colorizationresults} \label{sec:seresults}

\subsection{Main Results}
\label{subsec:mainRes}
{We compared LUCSS to LBIE \cite{Jianbo2018RAM} for scene sketch colorization, since 
to the best of our knowledge, LBIE is the only existing colorization work with the same goals as ours, i.e., taking a commented sketch as input, and outputting a color image. 
In general, the architecture of LBIE is similar to the RMI based architecture used for object  colorization of LUCSS. Both of these two architectures are able to colorize an entire input sketch in a {single step} by implicitly inferring the correspondence between language descriptions and objects (or object parts). We therefore also include the RMI based architecture shown in Figure~\ref{fig:instanceNetwork}, which takes as input a scene  sketch  and a description of scene-level colorization requirement, and generates a color image in a singe step instead of  two steps used by LUCSS, as another baseline method. In the following paragraphs of this section, these three comparison approaches are called LUCSS, LBIE, and sRMI for short.}

\subsubsection{Data Collection}  
\begin{figure}[bth]
\centering
   \includegraphics[width=0.99\linewidth]{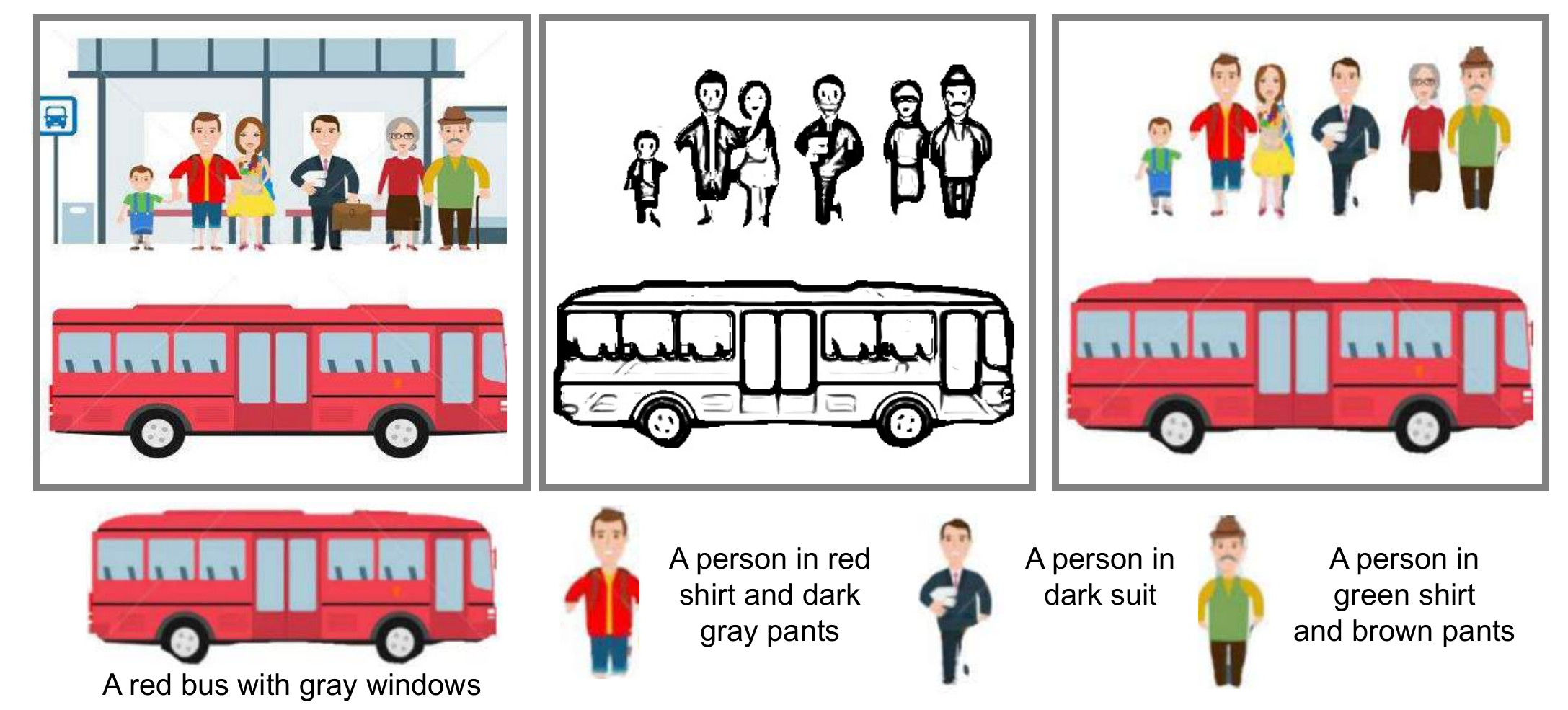}
  \vspace{-7pt}   
        \caption{Illustration of training data collection for object instance colorization. Top row: a reference image from SketchyScene (left); filter response of Hed \cite{Xie2015HED} and  X-DoG \cite{XDOG2011} (middle); cut-out color object instances (right). Bottom row: user-written descriptions of the representative color object instances. }
\label{fig:instanceData}
  \vspace{-7pt}   
\end{figure}

\paragraph{\textbf{Data collection for LUCSS: object colorization.} }
{We collected three modalities of data to train the models of object colorization:  color object instance, edge map, and text description (caption). We extracted color object instances and their edge maps from the 5,800 reference cartoon style images of SketchyScene.} Figure \ref{fig:instanceData} illustrates how the training data was prepared. 
We first leveraged Mask-RCNN trained on MS COCO to detect color object instances and then cut them out of the reference images. 
For each instance, we fused the filter responses of Hed \cite{Xie2015HED} and X-DoG \cite{XDOG2011} as the training edge map (object sketch). 
In total we collected 3,739 sets of object examples, each set consisting of a caption authored by crowd workers, a 192 $\times$ 192 color image, and a 192 $\times$ 192 edge map. 
The captions covered object instances from 20 categories, namely, moon, sun, cloud, house, bench, road, bus, car, bird, people, butterfly, cat, chicken, cow, dog, duck, sheep, tree, rabbit, and pig, which describes object  instances in 15 different colors. {The number of colors varies from one to three for each individual object instance (e.g.,  there are two colors for a red  bus with gray windows). We further split the 3,739 sets of examples into two parts: 2,814 sets of examples used for training data, 357 sets for validation, and 568 sets for test.}

{
\textbf{Data collection for LUCSS: background colorization.}} 
{For background colorization, we used the following three modalities of training data: 
cartoon style color images, their captions, and color foreground object instances with blank background (See the leftmost image of Figure \ref{fig:bg_lbieRmi} for an illustration).}
We used two types of strategies to extract the training data from the reference images of SketchyScene. The first one cut out objects from the reference images with mask-RCNN, and changed the color of all the pixels outside of the objects white. 
This strategy did not work well on some reference images where some object instances could not be detected. For these reference images, we used the other strategy: using the boundary of an object instance in the scene sketch as the boundary of the corresponding color object in the corresponding reference image (the object correspondence can be obtained from SketchyScene). We employed 5 workers to generate the captions. For each set of examples, the color details of two / three components, ``sky", ``land", and/or ``background", were described. In total, we collected 1,328 sets of training examples. We used 1,200 sets of examples for training,  the remaining 128 sets were equally split for validation and test. 
All the images were resized to $768 \times 768$ pixels.

\textbf{{Data collection for sRMI and LBIE.}} Similar to the data collection for LUCSS, we collected three types of data for sRMI and LBIE: 
(1) color images selected from the reference images of SketchyScene ({we selected the reference images which were in the training set of SketchyScene and had the majority pixels belonging to the above-mentioned 20 categories}), 
(2) corresponding edge maps, 
and (3) corresponding captions. 
To collect the captions in scales, we developed an on-line system to assist the crowd workers in generating captions for cartoon scenes. To ensure the description style is close to that generated by the captioning algorithm, we used a template-based approach for caption generation.
{More specifically, each sentence of a caption comes from a repository of caption templates, which describes the instances and their corresponding quantities, colors, and spatial relationship as Algorithm~1. }
In this way, We employed 24 workers and collected 1,328 sets of examples, 1,200 of which were used for the training and the rest  for validation.


\textbf{{Test data.}}  {We selected 100 scene sketches from the test set of SketchyScene as the test data ( the other scene sketches do not depict reasonable scenes if the objects
outside the selected 20 categories are removed from the scene sketch ). Each scene sketch we produced had up to three captions (varying on different evaluation tasks). Apart from being used to evaluate the colorization performance of LUCSS, LBIE, and sRMI, this test data has also been used for the experiments on the analysis of the components of LUCSS.}

\subsubsection{Experimental Settings} 
For object instance colorization, we used a batch size of 2 and trained with  $100K$ iterations. 
In the initial phrase of the training, we used the ADAM optimizer \cite{KingmaB14} and set the learning rate of generator at $0.0002$ and that of discriminator at $0.0001$.  After $50K$ iterations, we adjusted the learning rate of discriminator to $0.0002$.
For background colorization, we used batch size of 1 and trained with $100K$ iterations. We set the initial learning rate for both the generator and discriminator at $0.0002$ and reduced it by $75\%$ after each $20K$ iterations. We set the iteration number of LSTM at 15, the cell size of mLSTM at 512 for both object instance and background 
colorization modules.

\textbf{sRMI and LBIE.} We trained $100K$ iterations for both sRMI and LBIE. We set the initial learning rate of both generator and discriminator at $0.0002$ and reduce it by $75\%$ after each $20K$ iterations for sRMI. We set the iteration number of LSTM at 200, the cell size of mLSTM at 128 for sRMI. The iteration numbers of LSTM in LBIE were adapted to the input captions. We set other experimental settings of LBIE by following \cite{Jianbo2018RAM}.  

\subsubsection{Comparison Results}
\vbox{}
{We conducted two sets of experiments to evaluate the performances of LUCSS, sRMI, and LBIE.}

\begin{figure}[tbh]
\centering
   \includegraphics[width=1.05\linewidth]{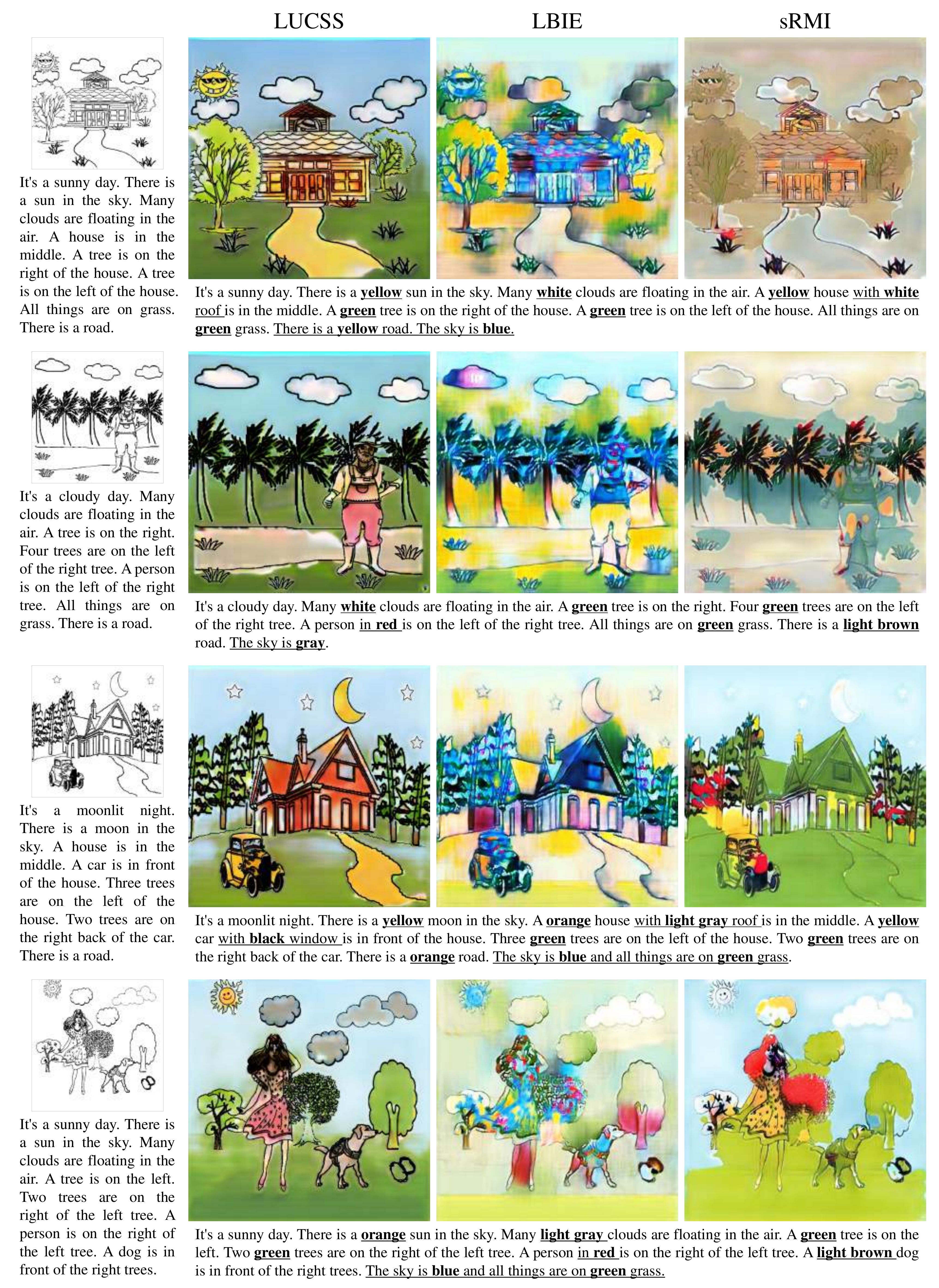}
  \vspace{-7pt}   
   \caption{LUCSS \textit{vs.} LBIE \textit{vs.} sRMI on various inputs. An input scene sketch and the automatically generated caption are shown on the left for each set of example. User edited captions are shown at the bottom of each colorization result. Each row shows the results produced by the three competitors. More results can be found in supplementary materials.} 
\label{fig:overlucss}
  \vspace{-7pt}   
\end{figure}

{\textbf{Single caption.}} In the first set of experiments, we compared LUCSS, LBIE, and sRMI on the {100} scene sketches of the whole test data. For each test scene sketch, we produced one caption (by re-editing the results of captioning ).  Figure~\ref{fig:overlucss} shows the visual results of these three competitors on some representative test examples. 
 
 Generally, LUCSS outperformed both LBIE and sRMI overwhelmingly. LUCSS colorized both objects (e.g., the roads, houses, clouds, moons, trees in the results), and backgrounds (the green land and sky) with smooth colors, while LBIE and sRMI generated a lot of color regions which covered more than one objects. sRMI relatively performed better than LBIE in some cases. We take the results in the third row of Figure~\ref{fig:overlucss} for example. sRMI colorized the sky and grass with smooth blue and green respectively, while the color generated by LBIE was mixed with multiple colors (e.g. white, yellow and pink).


With respect to {\em faithfulness}, which measures whether the colorization follows the language instructions, LUCSS achieved significantly better performance than LBIE and sRMI as well. We take the example in the fourth row of Figure~\ref{fig:overlucss} to explain our observation. LUCSS colorized all the clouds light gray, while LBIE colorized them into different colors including white and green, and sRMI colorized them into white, green and orange; LUCSS painted the lady red following the caption, while LBIE painted her into blue and red gradient, and sRMI colorized her with red hair and orange skirt. Relatively, sRMI performed a little better than LBIE on some examples. For example, sRMI can colorize the sky and grass in the third row of Figure~\ref{fig:overlucss} correctly, while LBIE failed to meet the demand of the caption.
 
The poor results of LBIE and sRMI may be mainly due to two factors. First, both LBIE and sRMI did not achieve accurate instance segmentation, which can be indicated by the tree at the most right side (LBIE: white and red; sRMI: green and blue), and the person (LBIE: blue skirt and red legs; sRMI: red hair, orange skirt and green legs). In contrast, LUCSS achieved accurate instance segmentation, since the boundaries of the resulting color objects were consistent with those in the sketch. Second, either LBIE or sRMI did not achieve accurate correspondence between the description and the objects. We can see the evidence from the sun (LBIE: blue sun; sRMI: white). In LUCSS, the correspondence is naturally obtained by the captioning module.
%
%


\begin{figure*}[tbh]
\centering
   \includegraphics[width=0.99\linewidth]{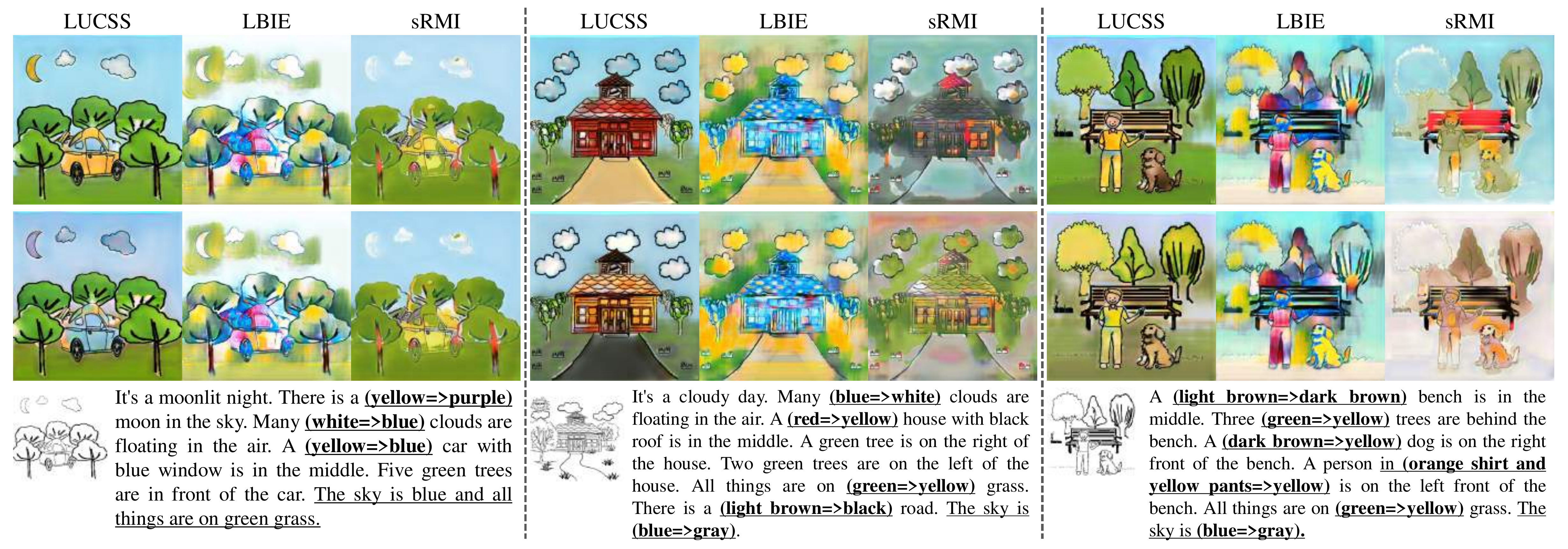}
  \vspace{-7pt}   
   \caption{Colorization results by LUCSS, LBIE, and sRMI with two  different sets of user-specified colorization requirements for each scene sketch. For each set of comparison, an input scene sketch is  at the bottom, with the corresponding caption. Text without underline is automatically generated, while underlined text is added by users. Arrows in the parentheses represent the change of colorization requirements (text on the left side of the arrows correspond to the results in the top row; text on the right side of the arrows correspond to the results in the bottom row).} 
\label{fig:multicaption}
  \vspace{-7pt}   
\end{figure*}

\textbf{Multiple captions}.
In the second set of experiment, we randomly selected 30 examples from the test dataset and produced three captions for each scene sketch. We then evaluated LUCSS, LBIE, and sRMI on all the three captions for each scene sketch. Some representative results are shown in Figure \ref{fig:multicaption}. The results of LUCSS generally met the user-specified colorization requirements. The results of LBIE on three different captions almost stayed the same. Although sRMI can response to the changed captions in some cases, it colored most objects in wrong colors. It indicate that both LBIE and sRMI failed to obtain the correspondence between the captions and objects. In \cite{Jianbo2018RAM}, LBIE showed the ability to learn the correspondence between words and objects, which is possibly because the captions in \cite{Jianbo2018RAM} contain much shorter sentences than these in this study (captions in our study typically contain more than 6 sentences, much longer than two or three sentences in \cite{Jianbo2018RAM}).


%% file: 6_result_object.tex
\subsection{Ablation Experiments}
In this section, we design various experiments to analyze the two major components of LUCSS. 
\subsubsection{Object Colorization}
\paragraph{\textbf{LBIE} versus \textbf{MRU-RMI}} In this set of experiment, we evaluated the performance of two types of models,  LBIE and MRU-RMI, for object colorization. The test data contained 568 sets of examples as discussed in the data collection section above. Each set of test examples contained an object sketch and a corresponding caption. LBIE is the same network as that used in the previous experiments. MRU-RMI is also the same network as sRMI,  which was used to colorize a whole scene sketch in the previous experiments.   

\begin{figure}[tbh]
\centering
   \includegraphics[width=0.99\linewidth]{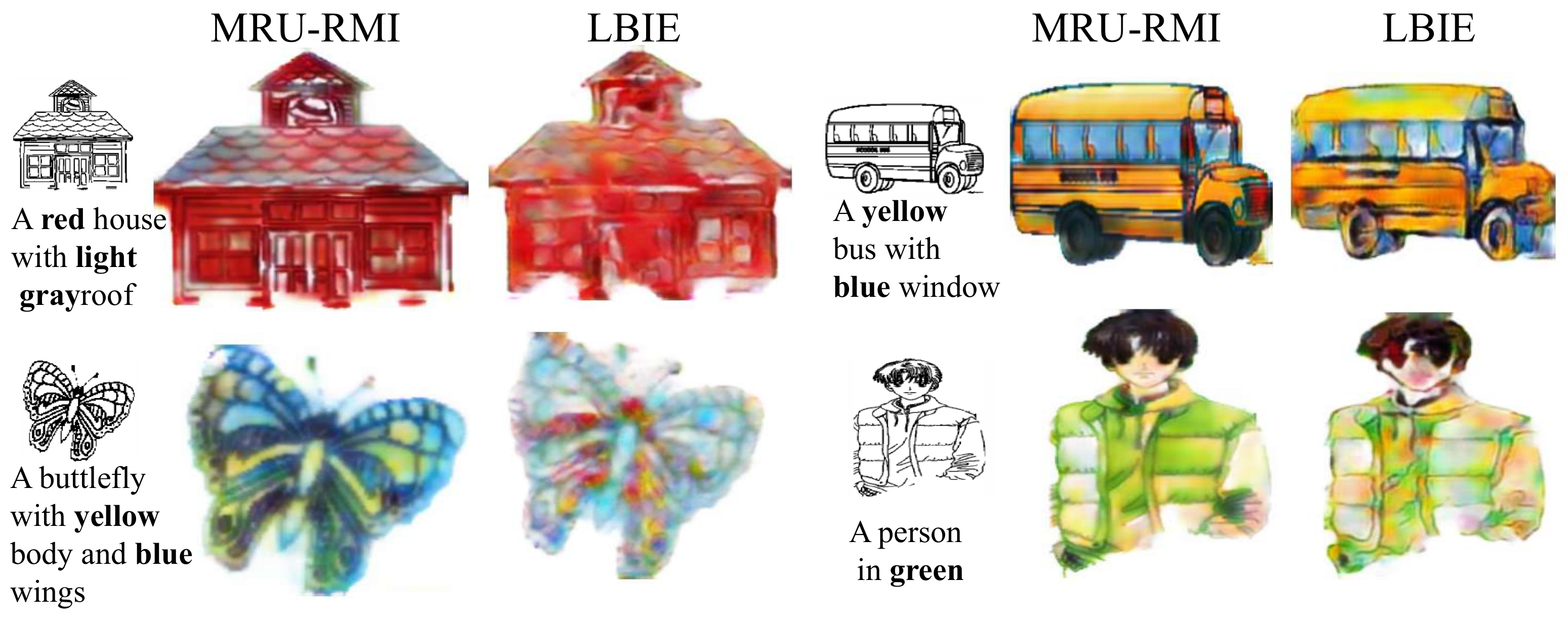}
  \vspace{-7pt}   
   \caption{Network architecture evaluation for object  colorization: MRU-RMI versus LBIE. MRU-RMI (shown in Figure~\ref{fig:instanceNetwork}) denotes the architecture used by LUCSS for object colorization. It uses MRU backbone based encoder \& decoder and RMI based fusion model. LBIE is the model used in \cite{Jianbo2018RAM}. Input scene sketches and language descriptions are shown on the left of each example. } 
\label{fig:obj_lbieMRU}
  \vspace{-7pt}   
\end{figure}

Figure \ref{fig:obj_lbieMRU} shows some representative results. It can be seen that both LBIE and MRU-RMI can learn an implicit part-level segmentation and colorize the object parts with the colors instructed by the captions (e.g., both LBIE and MRU-RMI assigned blue to the windows of the bus). Relatively, 
MRU-RMI achieved better performance on inferring the correspondence between words in the caption and object parts. This can be told from the colorization results for the roof of the house and the windows of the bus.  
Combining the results in both Figure \ref{fig:obj_lbieMRU} and \ref{fig:multicaption}, 
we can see that MRU-RMI is powerful for object-level sketch images and its performance would be significantly degraded when the sketch image size goes up to a large-size complex scene sketch. Moreover, for the face region of the person in Figure \ref{fig:obj_lbieMRU}, we can see that MRU-based network also outperformed the atrous-convolution-based LBIE on the encoding of object-level sketch image features (performance of MRU based network is significantly degraded when the complexity and
the size of sketch image increase).
\paragraph{\textbf{MRU} versus \textbf{ResNet} blocks versus \textbf{Pix2Pix}} 
In this set of experiments,  we evaluated three different types of backbones for the encoder and decoder of the architecture used for object colorization.  
The first type of backbone is MRU, which is used by SkecthyGAN\cite{sketchGAN} as well as the encoder and decoder for object colorization of LUCSS.  ResNet denotes the residual block proposed by \cite{HeZRS16}. Pix2Pix denotes the convolution block used by the encoder and decoder of \cite{pix2pix2017}. For comparison purposes, we replaced the cascade blocks in Figure \ref{fig:instanceNetwork} with ResNet/Pix2Pix, and produced results on all the 568 sets of test examples. 
\begin{figure}[tbh]
\centering
   \includegraphics[width=0.99\linewidth]{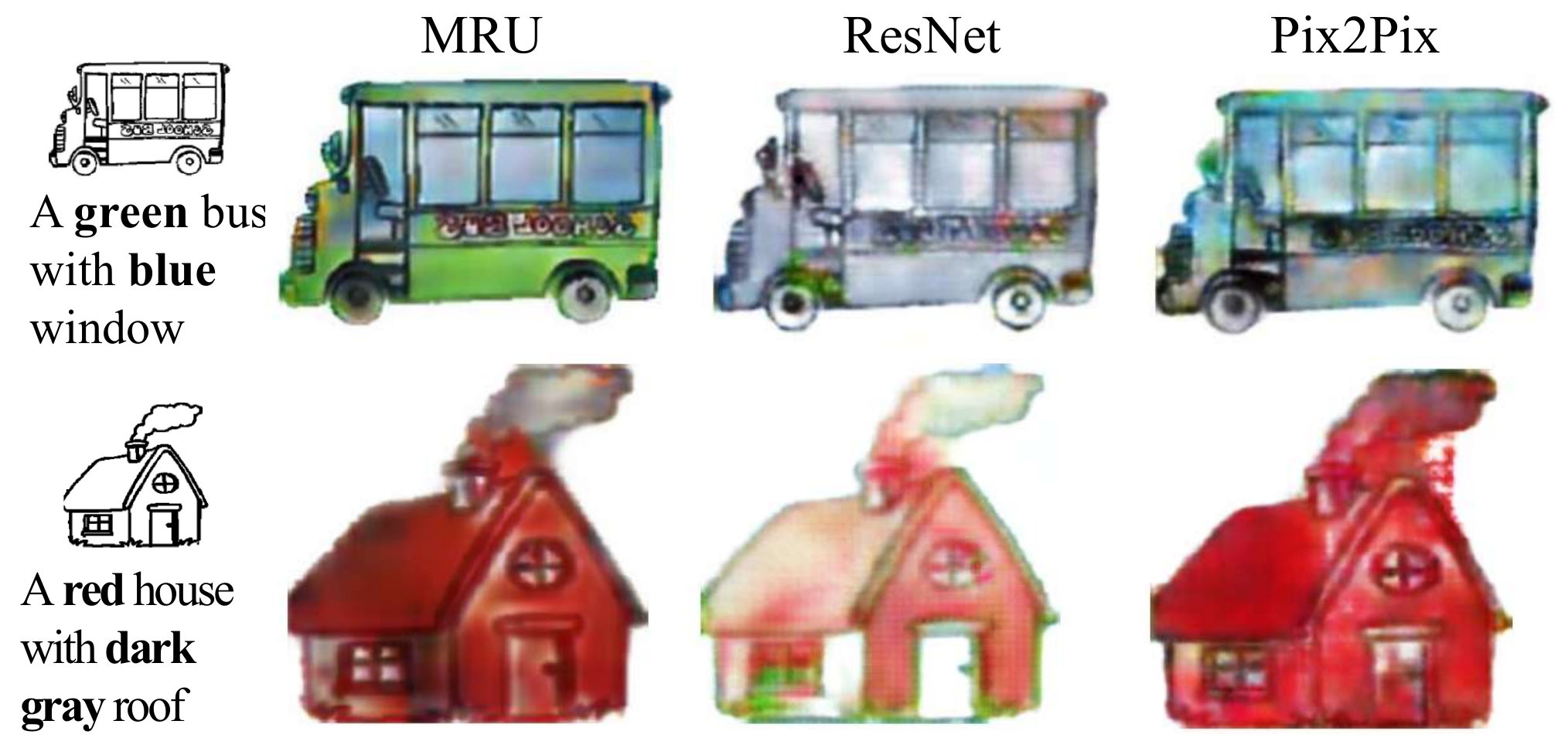}
  \vspace{-7pt}   
   \caption{Network backbone evaluation for object  colorization : MRU versus ResNet versus Pix2Pix. MRU based image encoder \& decoder outperforms both ResNet and Pix2Pix based image encoder \& decoder.} 
\label{fig:obj_backbone}
  \vspace{-7pt}   
\end{figure}

In Figure \ref{fig:obj_backbone}, we show the results of two representative examples. MRU achieved better performance than both Pix2Pix and ResNet. Specifically, MRU generated clearer texture and object part boundaries compared to ResNet and Pix2Pix (e.g., see the window and the body of the colorized bus).  In the aspect of following language instruction, the results of MRU are also superior to those of ResNet and Pix2Pix. This also indicates that the whole network based on MRU can infer more accurate correspondence between the caption and image content using the image features extracted by MRU.

%% file: 6_result_background.tex
\subsubsection{Background Colorization}

In this section, we first investigate the architecture of the GAN for 
the background colorization task, and then study what kind of backbone is more appropriate for this task.

\paragraph{\textbf{LBIE} versus \textbf{ResNet-RMI}} 
In this set of experiments, we compared two types of architectures, which can be used for background colorization. The first architecture is LBIE. The other is the architecture discussed in Section \ref{sec:bg_color}. We call this architecture ResNet-RMI in this section since the backbone of the generator uses ResNet blocks.  Both LBIE and ResNet-RMI were trained with 1,200 sketches in the training dataset, and tested
on the test dataset. 
%
%

\begin{figure}[tbh]
\centering
   \includegraphics[width=0.99\linewidth]{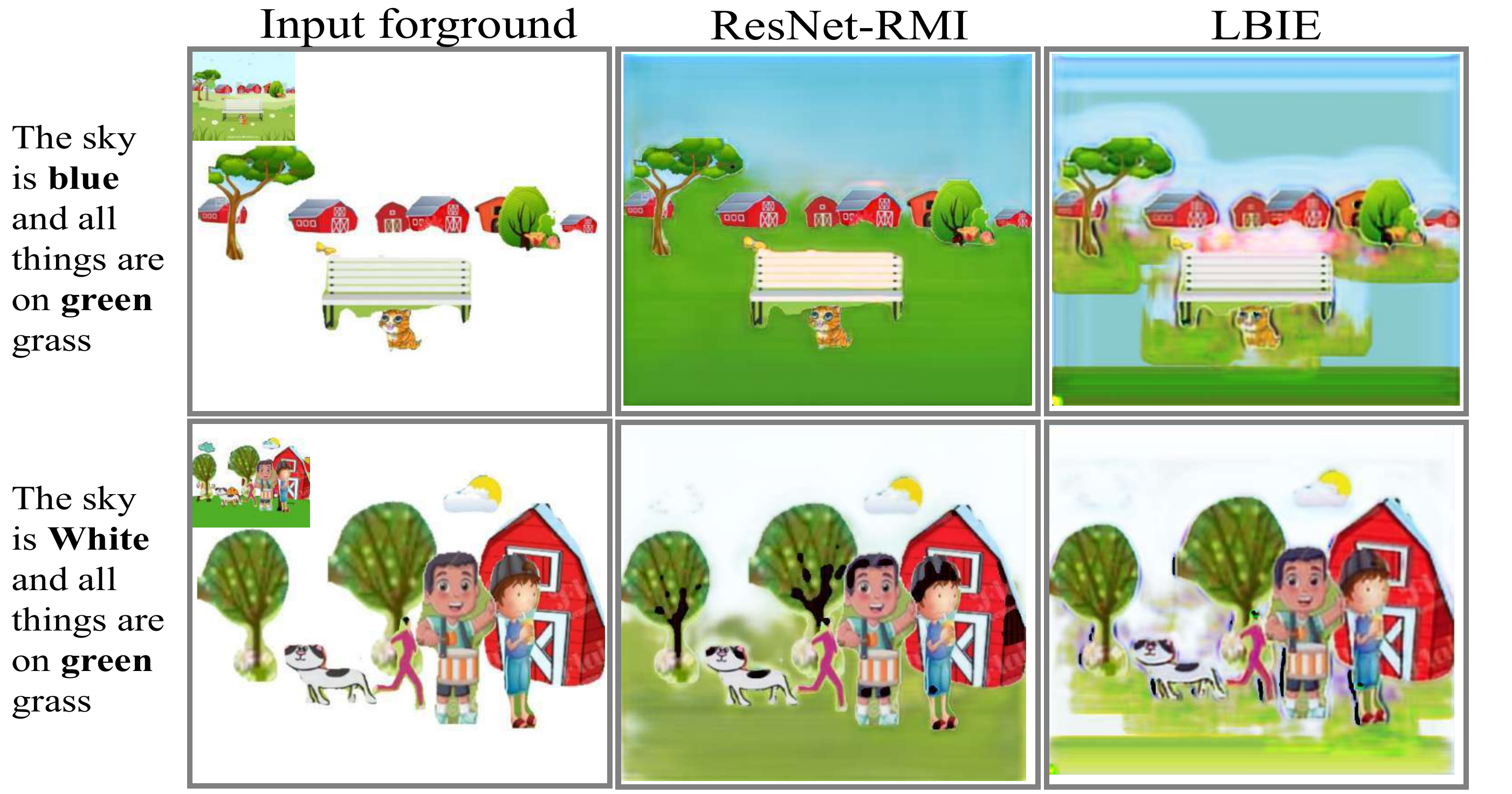}
  \vspace{-7pt}   
   \caption{Network architecture evaluation for background colorization : ResNet-RMI versus LBIE. 
 It took as input the language descriptions on the left, and images containing color foreground objects and blank backgrounds.  The reference images from which color foreground objects were cut out are shown on the top left corner of each input image.} 
\label{fig:bg_lbieRmi}
  \vspace{-7pt}   
\end{figure}

In Figure \ref{fig:bg_lbieRmi}, we illustrate the results of  LBIE and ResNet-RMI on two sets of representative examples.  Both LBIE and ResNet-RMI successfully colorized the regions of sky and grass with the colors specified in the captions.  This indicates that the encoding modules of image, text and feature fusion modules of LBIE and ResNet-RMI work well. However, the results by ResNet-RMI were visually more pleasing than LBIE. It may be caused by the fact  that LBIE uses an asymmetrical  encoder-decoder architecture (the encoder of LBIE uses atrous convolution while the decoder uses regular convolution). 
%
%

\begin{figure}[tbh]
\centering
   \includegraphics[width=0.99\linewidth]{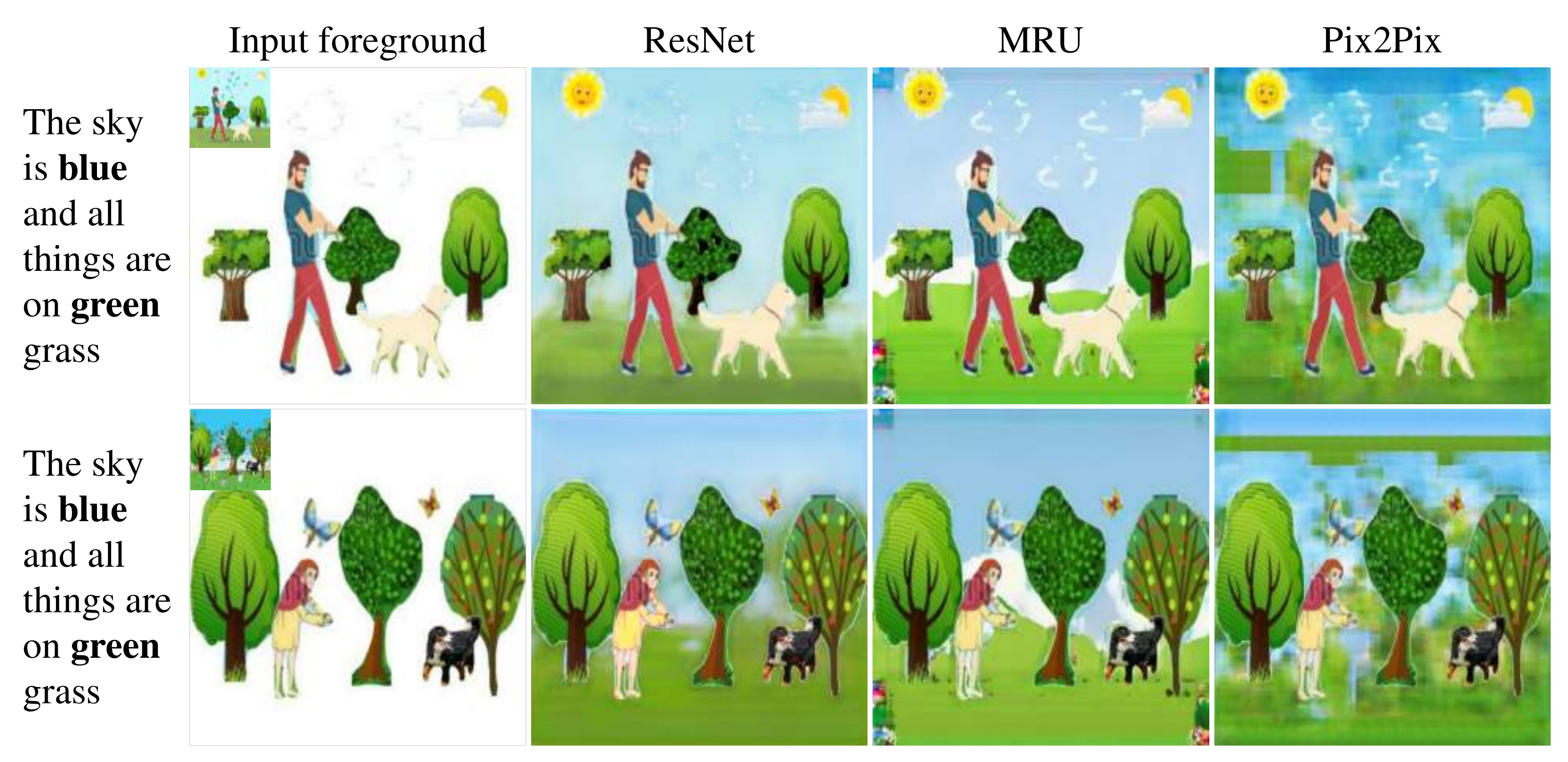}
  \vspace{-7pt}   
   \caption{Network backbone evaluation for background colorization : ResNet \textit{vs.} MRU \textit{vs.} Pix2Pix. ResNet outperforms both MRU and Pix2Pix when colorizing blank backgrounds surrounding color foreground objects.} 
\label{fig:bgBackbones}
  \vspace{-7pt}   
\end{figure}

\paragraph{\textbf{MRU} versus \textbf{ResNet} versus \textbf{Pix2Pix}} 
In this set of experiments, we study which backbone network is more suitable for large-scale image background colorization. We still used the three types of backbone network: MRU, ResNet, and Pix2Pix.  Results on representative examples are shown in Figure~\ref{fig:bgBackbones}.  We can see ResNet outperforms Pix2Pix in terms of visual effects, which can be explained by the fact that ResNet is much deeper than Pix2Pix (a ResNet block has three convolution layers while a  Pix2Pix block only has a single convolution layer). ResNet also has visually better results than MRU {(e.g., MRU generated some artifacts surrounding foreground objects, while ResNet doesn't)}, mainly because MRU is specialized for object-level sketches. When the MRU based network is used for a large size of color image, its performance is degraded (we can get the evidence by comparing  the results in Figure~\ref{fig:multicaption} and Figure \ref{fig:obj_lbieMRU}, additional evidence can be seen in SketchyGAN\cite{sketchGAN}).

%% file: 6_result_userscore.tex
\subsection{Human Evaluation}

To further justify the captioning and colorization results of LUCSS, we have designed and carried out two sets of user experiments: \textit{faithfulness study} and \textit{effectiveness study}. 
We recruited 11 participants for the studies.  All participants are undergraduates with no prior knowledge of this project. 

Each set of experiments covers the two sub-steps of  LUCSS: object colorization and background colorization. We did not conduct a study for system level performance  of LUCSS,  LBIE, and sRMI (see Section ~\ref{subsec:mainRes}) because LUCSS  obviously outperforms LBIE and LBIE outperforms sRMI on all the examples of the test dataset.

In the object colorization task, we randomly select 60 sketches from 20 classes as the input (3 sketches per class) and compare the results using two models: LBIE and MRU-RMI (our model).
In the background colorization task, we random select 20 scene sketches as the input and compare the results via two models: LBIE and ResNet-RMI (our model).
In both tasks, we further the performance of three backbones: MRU, ResNet blocks, and Pix2Pix.

In the \textit{faithfulness study}, our goal is to evaluate whether the colorization results are consistent with the text description. 
Each participant was given the input caption with the corresponding colorization results from one of the two models, or one of the three backbones. 
We asked the participant to pick out the colorization result which best fits the text description.

In the \textit{effectiveness study}, we compared the overall visual quality of the colorization results. 
Each participant was given the original sketch with the corresponding colorization results from one of the two models, or one of the three backbones. 
We asked the participants to pick out the most visually pleasing color image for the sketch.

\begin{figure}[tb]
\centering
   \includegraphics[width=0.95\linewidth]{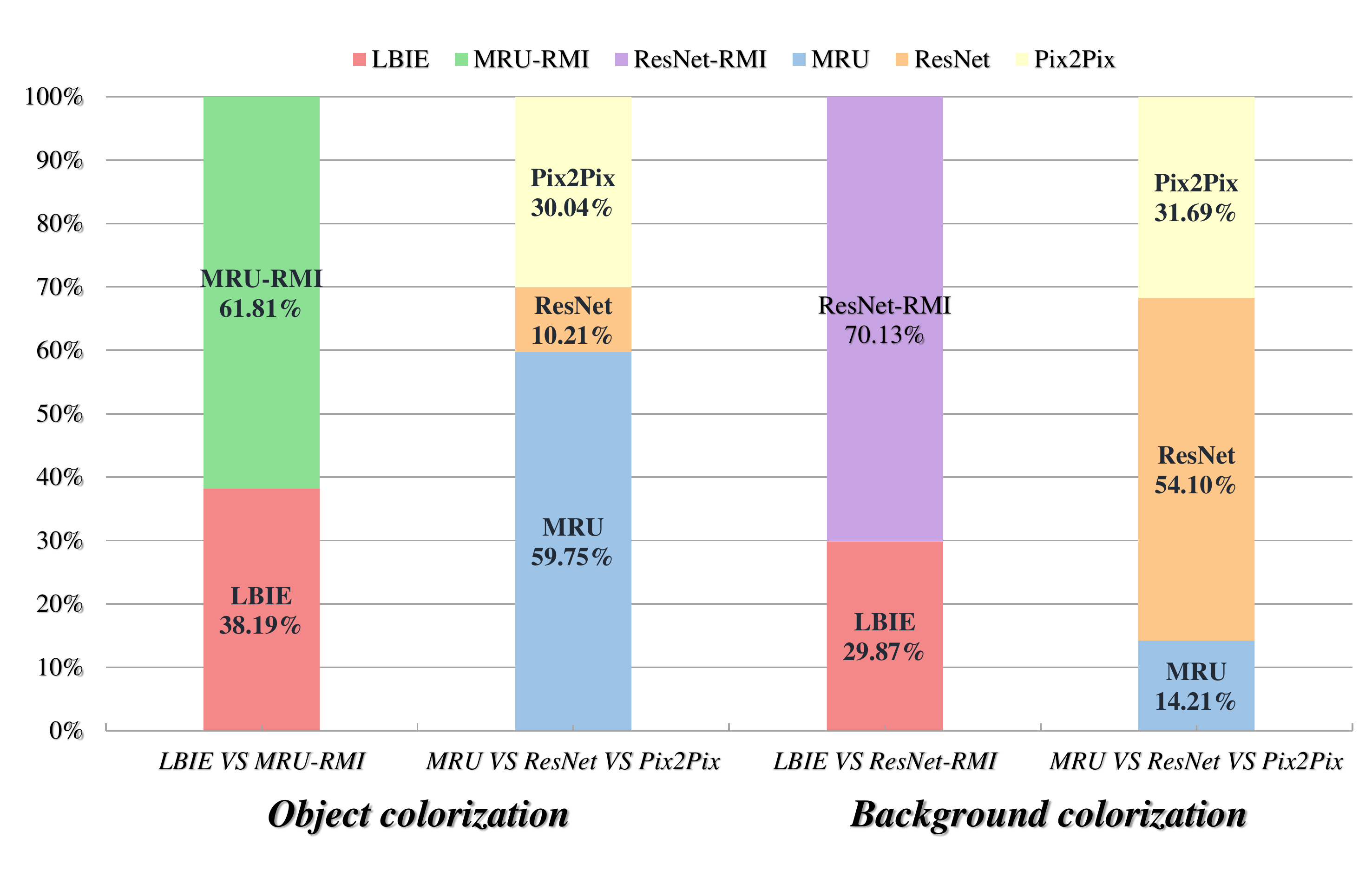}
  \vspace{-7pt}   
   \caption{Quantitative results for the faithfulness study. LUCSS uses the RMI model in general, MRU backbone for object colorization, and ResNet blocks for background colorization.}
\label{fig:faithfulness}
  \vspace{-7pt}   
\end{figure}

\begin{figure}[tb]
\centering
   \includegraphics[width=0.95\linewidth]{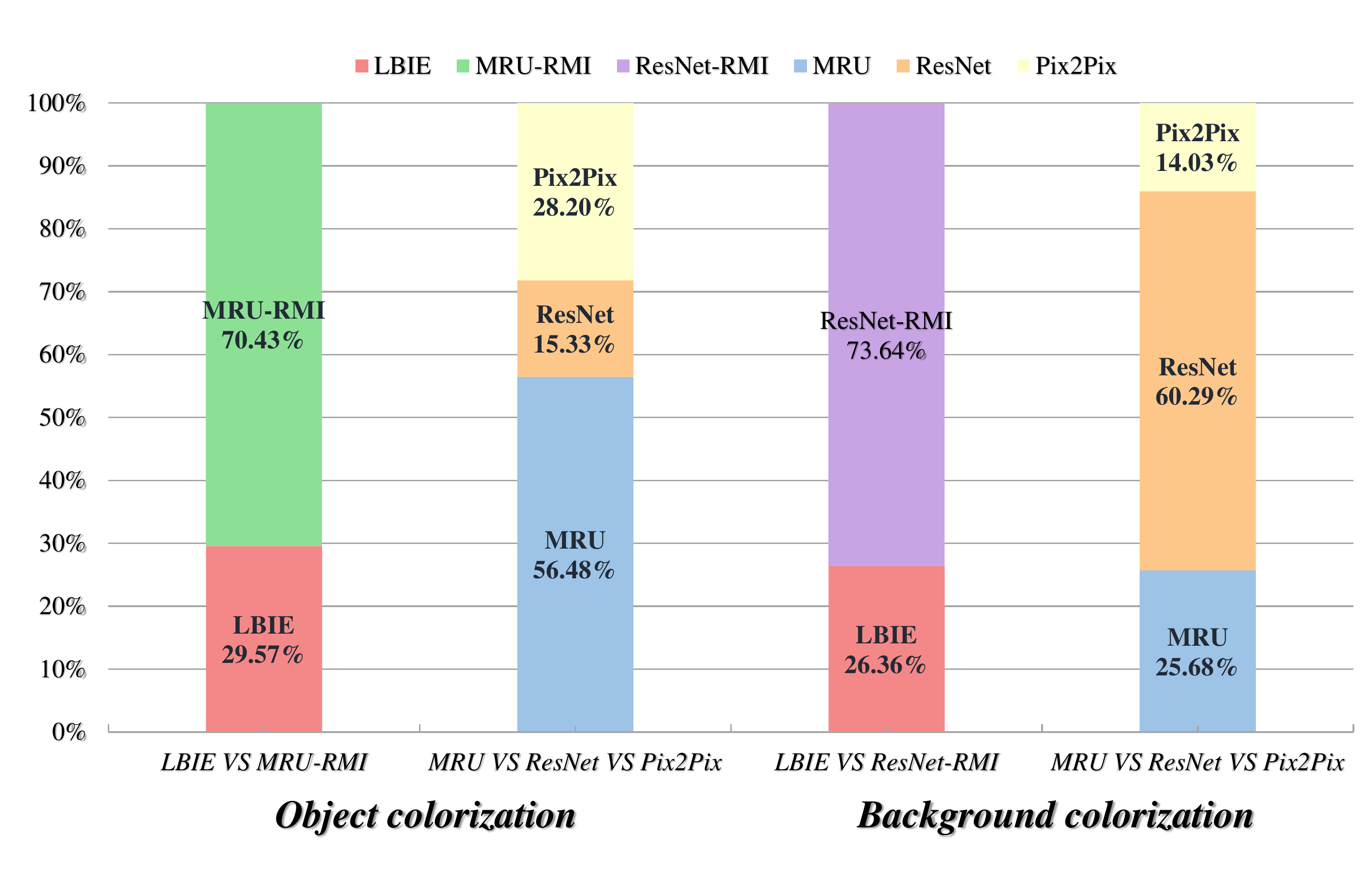}
  \vspace{-7pt}   
   \caption{Quantitative results for the effectiveness study. LUCSS uses the RMI model in general, MRU backbone for object colorization, and ResNet blocks for background colorization.}
\label{fig:effectiveness}
  \vspace{-7pt}   
\end{figure}

Overall, in each study, we have collected $11\times 20\times 3 = 660$ trials for the object colorization task and $11\times 20=220$ trials for the background colorization task. 
For both studies, we compare the selection rate of different models and backbones in Fig. \ref{fig:faithfulness} and \ref{fig:effectiveness}. 
For both tasks, our MRU-RMI model and ResNet-RMI model greatly outperform the LBIE model in user evaluation. 
Based on our user evaluation, LUCSS is more faithful and produces more visually pleasing results to users.
For object colorization, the MRU backbone stands out; while for background colorization, the ResNet backbone outperforms the rest.

%% file: 7_discuss.tex
\section{Conclusion, Discussion, and Future Work}
Understanding low-level scene sketches containing multiple sketched objects is a rarely studied problem. This problem is very challenging, especially when objects occlude each other in the depicted scene. The sparse nature of scene sketches leads to inferior performance of current models, even advanced ones, that are designed for natural images. In this work, we proposed a system called LUCSS to study how we can make machines understand complex scene sketches by adapting the existing powerful deep models for natural images to sketches, as well as how this understanding can benefit related applications. We have focused language-based interactive colorization of scene sketches.  
 
The current performance of LUCSS is limited by the segmentation performance since the highest mask AP is near $60\%$. It is necessary to propose more powerful algorithms by incorporating the characteristics of scene sketches with advanced models. Another limitation is that our current human-computer interaction needs improvement, though it is currently a practical solution considering the challenges of instance segmentation and inferring accurate correspondence between language descriptions and objects in the scene. As these challenging problems are solved by new datasets and more powerful models, LUCSS has the potential to become a far more mature system that could potentially generate fine-grained sketch colorization like generating a boy with ruddy cheek, or animating objects in a scene sketch by human speech commands. 

Image synthesis often has problems when data is left out. Since LUCSS is limited by training data, LUCSS generates cartoon images rather than natural images. Complex sentences are input to LUCSS which then processes them, filling in any missing detail such as omitted colors, and outputs an image.

 
%
%
%
This raises a question: is it possible to synthesize user-customized scene-level natural images by adding user-drawn sketches to the images of MS COCO's dataset to corroborate the existing language descriptions? 

LUCSS has great application prospects in the field of child education, accessibility, and media production. For instance, by translating the auto-generated caption text to human voice, LUCSS has the potential to read scene sketches to children and the blind. Via the interactive colorization system, both children and adults could create their own cartoon story books. 
In addition, the involved techniques in LUCSS may be useful in CAD and Virtual Reality (VR) industries. With text (voice) commands, LUCSS unlocks the potential to easily change the color schemes of a sketch scene and virtual environments.